\documentclass{article}

\usepackage[preprint]{neurips_2026}

\usepackage{amsmath,amsfonts,bm}

\def\eqref#1{equation~\ref{#1}}

\def\1{\bm{1}}

\DeclareMathAlphabet{\mathsfit}{\encodingdefault}{\sfdefault}{m}{sl}
\SetMathAlphabet{\mathsfit}{bold}{\encodingdefault}{\sfdefault}{bx}{n}

\usepackage[utf8]{inputenc}
\usepackage[T1]{fontenc}
\usepackage{hyperref}
\usepackage{url}
\usepackage{xcolor}
\usepackage{microtype}

\usepackage{amsmath,amssymb,amsfonts,mathtools}

\usepackage{graphicx}
\usepackage{booktabs}
\usepackage{multirow}
\usepackage{tabularx}
\usepackage{colortbl}
\usepackage{caption}
\usepackage{subcaption}
\usepackage{wrapfig}
\usepackage{sidecap}
\usepackage{float}

\usepackage{enumitem}
\usepackage{cleveref}
\usepackage{etoc}
\usepackage{titletoc}
\usepackage{xspace}
\usepackage{soul}

\usepackage{tcolorbox}
\tcbuselibrary{skins,breakable,listings}

\definecolor{citationblue}{HTML}{4b7a57}
\hypersetup{
    colorlinks=true,
    citecolor=citationblue,
    linkcolor=black,
    urlcolor=citationblue
}

\sidecaptionvpos{figure}{t}

\definecolor{pairwisecol}{HTML}{2a78d6}
\definecolor{directcol}{HTML}{1baf7a}

\tcbset{
    exbase/.style={
        enhanced,
        breakable,
        colback=white,
        colframe=black!30,
        colbacktitle=black!6,
        coltitle=black,
        fonttitle=\small\bfseries,
        fontupper=\small,
        fontlower=\small,
        boxrule=0.4pt,
        arc=1.5pt,
        left=5pt,
        right=5pt,
        top=3pt,
        bottom=3pt,
        before skip=6pt,
        after skip=6pt,
        segmentation style={black!30, dashed},
        before upper={\setlength{\parskip}{3pt}},
        before lower={\setlength{\parskip}{3pt}}
    }
}

\newtcolorbox{exbox}[2][]{
    exbase,
    title={#2},
    colupper=black!70,
    #1
}

\newtcblisting{promptbox}[2][]{
    exbase,
    unbreakable,
    listing only,
    title={#2},
    listing options={
        basicstyle=\ttfamily\scriptsize,
        breaklines=true,
        columns=fullflexible,
        keepspaces=true,
        aboveskip=0pt,
        belowskip=0pt
    },
    #1
}

\newcommand{\boxlabel}[1]{%
    {\scriptsize\sffamily\bfseries\MakeUppercase{#1}}\par\smallskip
}

\newcommand{\science}{\textsc{Science}\xspace}
\newcommand{\maths}{\textsc{Maths}\xspace}
\newcommand{\code}{\textsc{Coding}\xspace}
\newcommand{\cyber}{\textsc{Cyber}\xspace}

\makeatletter
\renewcommand{\@noticestring}{Preprint. Code: \url{https://github.com/Mamiglia/scalable-capability-evaluation}}
\makeatother

\title{Evaluating Model Capabilities After Benchmark Saturation}

\author{
  Morio Hamada$^{1*}$ \qquad Jehyeok Yeon$^{1,2,3,4*}$ \qquad Matteo Migliarini$^{1,5*}$ \\
  {\bfseries Justin Olive$^{6}$ \qquad Magda Dubois$^{7}$} \\[0.8em]
  {\normalfont\small $^{1}$LASR Labs \quad $^{2}$ELLIS Institute Tübingen \quad $^{3}$Max Planck Institute for Intelligent Systems} \\
  {\normalfont\small $^{4}$Tübingen AI Center \quad $^{5}$Sapienza University of Rome \quad $^{6}$Generality Labs \quad $^{7}$UK AISI} \\[0.3em]
  {\normalfont\small $^{*}$Equal contribution}
}

\begin{document}

\maketitle

\begin{abstract}
Rapid advances in AI capabilities are causing benchmarks to saturate.
Once several models approach perfect scores, a benchmark loses its power to differentiate between them, and with it its utility for informing AI safety cases and frontier policies.
Yet benchmark items produce far more information than the scores we typically extract from them: a full transcript of \textit{how} each model solved the task.
We investigate whether these transcripts can be used to assess model capabilities even after benchmark scores have saturated.
We simulate saturation retrospectively using historical benchmark and model releases, and test whether protocols that access only correctly solved transcripts from saturated benchmarks can predict capabilities measured on held-out, unsaturated items.
We evaluate representative contemporary approaches for extracting information from model transcripts: pairwise judging, single-transcript absolute scoring, and predictions from deterministic transcript features or embeddings.
We find that saturated transcripts retain capability information beyond benchmark accuracy. Amongst our evaluated protocols, pairwise judging performs best, reducing prediction error by $5.3\times$ compared to a baseline using only saturated benchmark scores.
This capability-relevant information is distributed across transcript sections, but concentrated in specific channels of information such as reasoning and tool-calls.
However, the predictive power of protocols does not always generalise across distribution shifts -- for example by underestimating the capabilities of the first reasoning models introduced in early 2025. Saturated benchmarks thus retain capability-predicting signal in their task transcripts, but extending the lifetime of existing benchmarks requires methods designed to extract these signals robustly.
\end{abstract}

\section{Introduction}
Benchmarks use a set of tasks and an evaluation protocol to estimate what an artificial intelligence system can do within a specific domain, enabling researchers to track the rate of progress. They play an important role in frontier AI safety, where evaluations help identify systems that may require stronger safeguards and inform governance decisions in domains such as biochemical applications, cybersecurity, and autonomous research \citep{openai2025preparedness,anthropic2026rsp,deepmind2026fsf}.

As models improve, however, their scores eventually converge near the ceiling~\citep{bowman2021fix,vania2021comparing}. At this point, the benchmark is \emph{saturated} and can no longer distinguish between models.
Benchmark saturation has already been shown to affect current evaluations, with a recent study finding signs of saturation on nearly half of 60 widespread LLM benchmarks~\citep{akhtar2026saturation}.

Current approaches to benchmark saturation largely rely on constructing harder benchmarks or continually updating existing ones. Researchers have introduced benchmarks such as Humanity's Last Exam and MMLU-Pro after earlier evaluations became too easy for frontier models \citep{phan2026hle,wang2024mmlupro}, while dynamic and continuously updated benchmarks aim to maintain discriminative power as models improve \citep{kiela2021dynabench,white2025livebench,jain2024livecodebench,ying2024updates}.
However, the time and cost required for constructing and verifying tasks make it difficult for continual benchmark updates to match frontier model progress \citep{ying2024updates}.
This creates a need to extract more information from existing benchmarks.

\begin{figure*}[t]
  \centering
  \includegraphics[width=.95\textwidth]{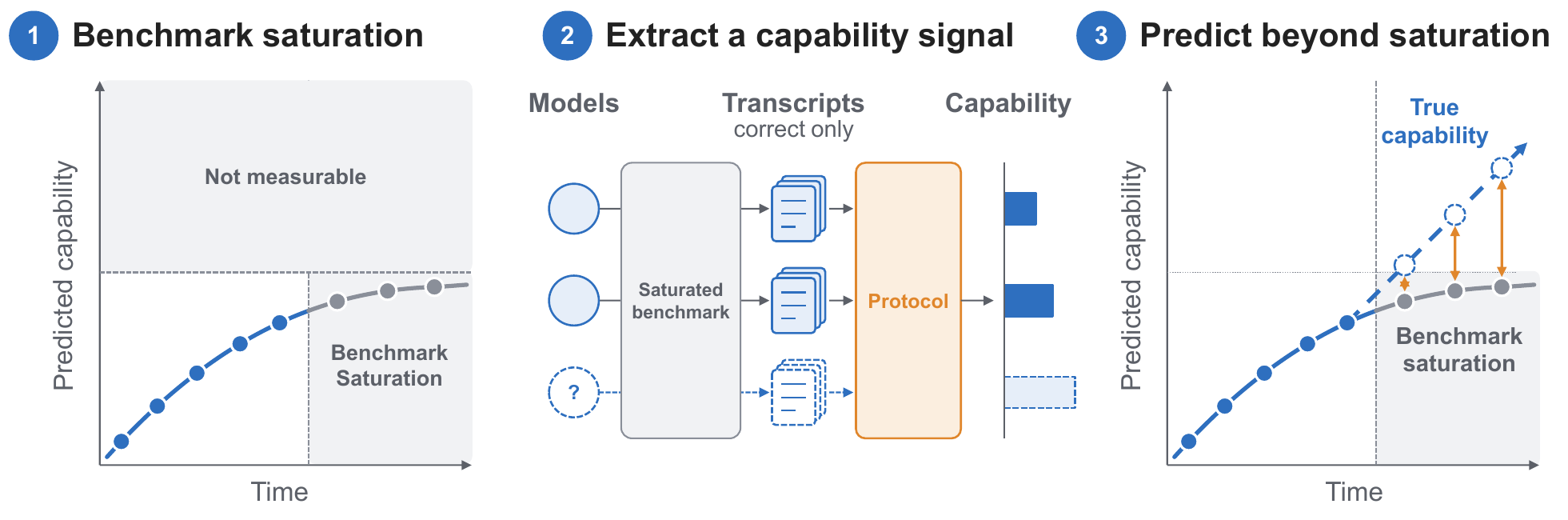}
  \caption{
  \textbf{Overview of our evaluation setting.}
  As benchmark scores saturate, it becomes infeasible to distinguish the capabilities of models.
  Using earlier models as references, protocols predict target model capability on harder, unsaturated items, using as information \textit{how} the models solve the tasks.
  }
  \label{fig:intro_fig}
  \vspace{-1em}
\end{figure*}

Benchmark scores compress model behaviour into a single scalar value, losing much of the information contained in evaluation transcripts. Models that appear identical under accuracy may differ in how they solve the same items.
Indeed, prior work has shown that solutions reaching the same correct answer can differ substantially in their intermediate reasoning~\citep{uesato2022solving,lightman2024lets}, and that transcripts can reveal intermediate progress~\citep{kapoor2026hal} and differences in solution quality~\citep{kirgis2026logs} that are not reflected in pass/fail outcomes.
Whether such transcript-level signals can be used to estimate capability once a benchmark has saturated, however, remains untested.
We therefore study whether transcript-level differences on saturated items retain capability signals that can be used to predict model capability on harder tasks beyond the benchmark's saturated scores (Figure~\ref{fig:intro_fig}).

We instantiate this question by assessing different methods for estimating capabilities from saturated benchmark transcripts. These methods were selected to represent contemporary approaches for extracting information from model outputs: deterministic transcript features \citep{bitton2025fingerprints}, transcript embedding, single-transcript scoring \citep{liu2023geval,kim2024prometheus2}, and pairwise judge comparisons \citep{zheng2023judge}. We then evaluate whether these approaches recover capability information once benchmark accuracy no longer reliably distinguishes between models.

Our main findings are as follows:
\begin{enumerate}[leftmargin=*,label=\arabic*.]
\item \textbf{Transcripts from correctly solved saturated items retain capability information beyond benchmark accuracy.} Pairwise LLM judging recovers the strongest signal, reducing prediction error 5.27$\times$ compared to a benchmark accuracy-based baseline.
\item \textbf{Capability-relevant information is distributed across transcripts and channels.} Removing any single third of a transcript does not fully impair predictions, suggesting that useful capability information is spread throughout the response. However, different channels of information matter in different domains: hiding model reasoning increases prediction error by roughly 3--4$\times$ in coding, cybersecurity, and science benchmarks, while removing tool use increases prediction error by about 4$\times$ in cybersecurity but has minimal effect in coding.
\item \textbf{Current capability-estimation protocols are not robust to all distribution shifts.} For example, pairwise judges from late 2024 could not reliably estimate the capabilities of the first reasoning models introduced in early 2025. This suggests that protocols can fail when new models bring paradigm shifts.
\end{enumerate}

\section{Problem Setting}
\label{sec:problem-setting}

We consider a setting in which model capabilities improve faster than new benchmarks can be developed and validated.
When differences in benchmark scores can no longer be used to separate models of different capabilities, we consider the benchmark to be \emph{saturated}.

Let $\mathcal{M}_{\mathrm{ref}}$ denote a set of reference models for which capability estimates are available.
Evaluating each reference model $r \in \mathcal{M}_{\mathrm{ref}}$ produces a set of transcripts $\mathcal{T}_r$ that record its behaviour on the benchmark.
Together with each reference model's capability $\theta_r$, these transcripts form the reference dataset $\mathcal{D}_{\mathrm{ref}}$:
\[
\mathcal{D}_{\mathrm{ref}}
=
\{(\mathcal{T}_r,\theta_r)
: r \in \mathcal{M}_{\mathrm{ref}}\}.
\]
We then consider a set of target models $\mathcal{M}_{\mathrm{target}}$ whose capabilities are to be estimated. For each target model $m \in \mathcal{M}_{\mathrm{target}}$, we generate its set of transcripts $\mathcal{T}_m$ on the same saturated benchmark, but its capability $\theta_m$ is unknown.
The problem is to estimate the capability of each target model from the reference data and the target model's transcripts. A \emph{protocol} $\mathcal{P}$ therefore produces:
\[
\mathcal{P}
\left(
\mathcal{D}_{\mathrm{ref}},
\mathcal{T}_m
\right)
\rightarrow
\hat{\theta}_m,
\qquad
m \in \mathcal{M}_{\mathrm{target}}.
\]

\section{Experimental Methods}
\label{sec:methods}
We instantiate our setting retrospectively using historical model releases and benchmark results.
For each prediction task, reference models precede the target period.
We additionally split benchmark items into an observed \emph{saturated set} and a harder \emph{unsaturated set}. Protocols receive only transcripts from the saturated set, while the unsaturated set is used exclusively to measure the ground-truth target capability $\theta_m^*$ that the protocols attempt to predict.

\subsection{Evaluation Setup}
\begin{figure}
    \centering
    \vspace{-1em}
    \hspace{1em}
    \includegraphics[width=0.9\linewidth]{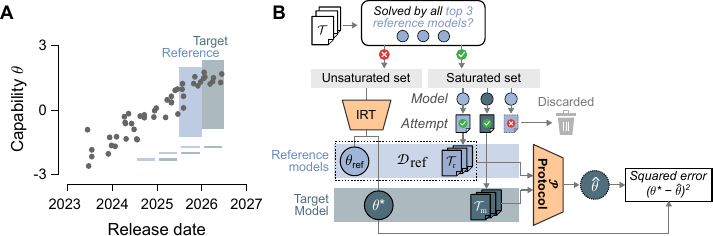}
    \caption{\textbf{Overview of the prediction task and evaluation methodology.}
    \textbf{(A)} Models are grouped into consecutive semesters; for each target period, the preceding semester serves as the reference set. \textbf{(B)} Items are split by whether the top-3 reference models answer correctly: easy items form the saturated set, of which only correctly solved transcripts $\mathcal{T}$ are provided to a capability-extraction protocol $\mathcal{P}$ to produce a prediction $\hat\theta$; the remaining items form the unsaturated set, on which an IRT model is fit to obtain the ground-truth capability $\theta^*$ for both references and targets.
    }
    \vspace{-0.5em}
    \label{fig:method_overview}
\end{figure}

\textbf{Reference and target periods.} We divide model releases into consecutive semesters. For each prediction task, the \emph{reference set} consists of models available in the preceding semester $\mathcal{M}_{t-1}$, while the \emph{target set} consists of models released during the subsequent semester $\mathcal{M}_{t}$. Our main evaluation uses two target periods: late 2025 and early 2026, with seven or eight reference models and eight target models in each period (Figure~\ref{fig:method_overview}A).

\textbf{Determining the saturation boundary.}  For each target period, we identify the three highest-capability models amongst the reference set.
We define an item as belonging to the \emph{saturated set} if all three models answer it correctly. The remaining items form the \emph{unsaturated set}.

\textbf{Prediction target.} To measure capability beyond the saturated portion of the benchmark, we use item response theory (IRT), which jointly estimates model capability and item difficulty from benchmark outcomes~\citep{baker1998item}. We fit a local IRT model to the reference and target models on the unsaturated set, while keeping each item's difficulty parameter $b_i$ fixed across semesters. The resulting capability estimate $\theta^{*}$ for each target model is used as the prediction target for the protocols. We provide the full IRT fitting procedure in Appendix~\ref{app:irt}.

\begin{table}[h]
    \small
    \centering
    \caption{
        \textbf{Evaluation domains and benchmarks.}
        Benchmarks used in each capability domain.
    }
    \label{tab:domains}
    \begin{tabularx}{\linewidth}{lX}
        \toprule
        \textbf{Domain} & \textbf{Benchmarks} \\
        \midrule
        Science &
        GPQA-Diamond \citep{rein2024gpqa},
        ARC-Challenge \citep{clark2018arc},
        HLE-Gold [Science] \citep{zhai2026hle}
        \\[0.3em]

        Mathematics &
        MMLU-Pro [Maths] \citep{wang2024mmlupro},
        AIME 2024 \citep{jia2024aime},
        AIME 2025,
        AIME 2026,
        APEX 2025,
        APEX Shortlist \citep{dekoninck2026matharena},
        HLE-Gold [Maths]
        \\[0.3em]

        Coding &
        AgentBench OS \citep{liu2024agentbench},
        SWE-Bench \citep{jimenez2024swebench},
        Terminal-Bench 2.1 \citep{merrill2026terminalbench}
        \\[0.3em]

        Cybersecurity &
        CyBench \citep{zhang2025cybench},
        InterCode \citep{yang2023intercodestandardizingbenchmarkinginteractive}
        \\
        \bottomrule
    \end{tabularx}
\end{table}

\textbf{Transcripts.} Our primary experiments test whether model capabilities can be inferred from \emph{how} models solve saturated tasks, rather than from whether they solve them correctly. We therefore provide each protocol only correctly solved transcripts from saturated items, thus removing correctness as a source of predictive signal. Unless otherwise stated, protocols receive the complete model response. The full framework is depicted in Figure~\ref{fig:method_overview}B.

\textbf{Domains.}
To assess generalisability, we evaluate four domains: \science, \maths, \code, and \cyber. Each domain combines benchmarks spanning a range of item difficulties, allowing us to distinguish between saturated and unsaturated items.
Table~\ref{tab:domains} provides a list of the benchmarks used to represent each domain.

\textbf{Model selection.} Our main experiments utilise the complete model output used during inference. For earlier models, this consists of the standard response transcript. For later reasoning models, however, reasoning traces are an important part of the transcript but are not consistently available through closed-source model providers. We therefore primarily use open-weight reasoning models in periods where reasoning models are included in the evaluation, while in earlier periods closed-source non-reasoning models are retained. We evaluate generalisation to closed-source reasoning models separately in Section~\ref{sec:results-generalization}. The full model list is provided in Appendix~\ref{app:evaluated-models}.

\textbf{Metrics.} For a protocol $\mathcal{P}$ and target model $m$, we measure prediction error using squared error:
$
\epsilon_m = (\hat{\theta}_m - \theta_m^*)^2,
$
where $\theta_m^*$ is the measured target capability and $\hat{\theta}_m$ is the prediction.

\subsection{Prediction Protocols}

We compare several contemporary approaches for extracting information from transcripts, and adapt them to estimate capabilities from the observations available in our setting. Table~\ref{tab:prediction_protocols} provides a summary of each of the protocols, while full implementation details are provided in Appendix~\ref{app:protocols}.

\begin{table}[h]
\centering
\caption{\textbf{Prediction protocols.} All non-baseline protocols extract transcript-level signals and calibrate them to target capability $\theta^*$ using the reference models.
}
\label{tab:prediction_protocols}
\begin{tabularx}{\linewidth}{l X}
\toprule
\textbf{Protocol} & \textbf{Signal extracted} \\
\midrule

Baseline protocol
&
--
\\

Transcript features
&
Deterministic properties (e.g., transcript length and prose features)
\\

Transcript embeddings
&
Text-embedding representation of a single transcript
\\

Absolute scoring
&
Absolute scalar score given to a single transcript
\\

Pairwise judging
&
Relative capability evidence from comparisons between two transcripts
\\

\bottomrule
\end{tabularx}
\end{table}

\textbf{Baseline protocol.}
The baseline represents the capability estimate available using benchmark scores alone under saturation.
Because the evaluated model solves all observed items correctly by construction, its accuracy provides no distinguishing signal.
Consequently, this baseline simply assigns every target model the mean capability of the reference models.

\textbf{Statistical tests.}
To assess the significance of a protocol's performance, we account for both prediction accuracy and uncertainty induced by the finite reference set through a comparison against the aforementioned baseline protocol (Section~\ref{sec:results-signal}). For each protocol, we estimate prediction uncertainty from leave-one-model-out errors on the reference models and use this to construct a Student $t$ distribution around each target-model prediction. Protocols are therefore penalised both for inaccurate predictions and for predictions that are unstable across reference models.

We compare each protocol with the baseline using the log-likelihood ratio (LLR) of the target model's observed capability under their respective predictive distributions. We then test whether the mean LLR across target models differs from zero using one-sample $t$-tests. Full details are provided in Appendix~\ref{app:prediction-stats}. For within-protocol comparisons in Sections~\ref{sec:results-source} and~\ref{sec:results-generalization}, we instead apply paired $t$-tests directly to target-model squared errors.

\section{Results}
To understand whether and how transcripts from saturated benchmarks can be used to estimate model capability, we ask three questions:
\textbf{(1)} Do correctly solved transcripts from saturated benchmarks contain any capability-predicting signal beyond raw accuracy?
\textbf{(2)} Where in the transcript is this signal located?
\textbf{(3)} Can we reliably extract this signal to predict capabilities across distribution shifts such as time or models?

\subsection{Saturated benchmark transcripts contain capability information}
\label{sec:results-signal}

\begin{wrapfigure}{R}{0.42\textwidth}
    \vspace{-1em}
    \centering
    \includegraphics[width=\linewidth]{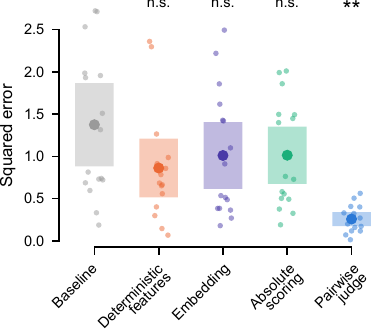}
    \caption{
        \textbf{Pairwise judges extract capability-predicting signal from saturated transcripts.}
        Prediction performance of each protocol for 16 target models  (squared error: lower is better). Each point is one target model; circle and box: mean and 95\% CI.
        **: $p<.01$, one-sample $t$-test of LLR against the baseline protocol; \textit{n.s.}: not significant.
        }
    \label{fig:signal-capability}
    \vspace{-1em}
\end{wrapfigure}

We first ask whether transcripts from saturated benchmark items contain information about capabilities. To do so, we assessed whether any of our protocols could recover a model's capabilities from only its correctly answered transcripts and a reference dataset (the capabilities and correctly answered transcripts of historical models). Figure~\ref{fig:signal-capability}
compares the prediction error of the four protocols over the 16 target models of 2025-II and 2026-I, each averaged over the four domains.

Pairwise judging predicted target capabilities most accurately (with the lowest error amongst our protocols: MSE=0.26, 95\% CI: [0.18, 0.34]). Its predictions also placed more likelihood on the targets' true capability than the baseline (mean LLR=0.68, one-sample $t$-test: $p=.005$;  see Appendix~\ref{app:results_signal}). This result was not specific to our choice of judge (Appendix~\ref{app:judge_choice}), and tended to improve with the number of items and reference models judged in most tasks (Appendices~\ref{app:item_abl} and~\ref{app:reference_item_abl}).

Other protocols were less reliable. While all tended to yield lower errors than the baseline (deterministic transcript features: 0.86 [0.52, 1.21]; transcript embeddings: 1.01 [0.61, 1.41]; absolute scoring: 1.01 [0.67, 1.35]), none placed significantly more likelihood on the true capability than the baseline.

Feature-based predictors did not perform significantly better than the baseline, suggesting that the capability signal is not primarily carried by simple properties such as response length or surface-level writing style. In contrast, pairwise judging substantially reduced the prediction error, indicating that it extracts information beyond such high-level features.

\subsection{Capability signals are distributed throughout transcripts}
\label{sec:results-source}

Having found
that saturated benchmark transcripts contain capability-relevant signals, we next ask \textit{where} this signal lies. This is practically important because model providers may not always make the full transcript available for closed-source models, and some protocols may have restrictions on how long a transcript they can judge efficiently. We focus on pairwise judging, the strongest protocol in Section~\ref{sec:results-signal}, and report the same analyses for the rest in Appendices~\ref{app:thirds-other-protocols} and~\ref{app:channels-other-protocols}.

\paragraph{Capability-relevant information persists across transcript sections.}
\label{sec:results-source-position}

\begin{figure}[!htbp]
    \centering
    \vspace{-.2em}
    \includegraphics[width=.82\textwidth]{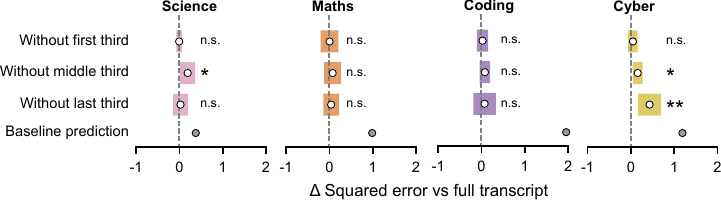}
    \caption{
    \textbf{Effect of removing transcript sections on prediction error.}
        Change in squared prediction error of pairwise judging when the first, middle, or last third of every transcript is removed, relative to the complete transcript (\textit{dashed line}). \textit{Grey points} show the baseline prediction. *: $p<.05$; **: $p<.01$.
    }
    \label{fig:thirds}
    \vspace{-1em}
\end{figure}

We first ask whether capability-relevant information is concentrated in particular parts of the transcript, which matters when limited context windows force protocols to observe only part of it. We divide each transcript into thirds and run two complementary tests: whether any third is necessary, by removing it and predicting from the remaining two, and whether any third is sufficient, by predicting from that third alone. When a third is removed, we measure the increase in prediction error relative to that obtained from the complete transcript; when a third is kept alone, we test whether it can still outperform the baseline protocol.

In \maths and \code, no single third was necessary for prediction. Removing a transcript section significantly increased squared error only in \cyber (middle third: $+0.16$, $p = .010$; last third: $+0.43$, $p = .003$), and less robustly in \science (middle third: $+0.19$, $p = .032$) (Figure~\ref{fig:thirds}) compared to the complete transcript. These increases were small compared to the gap between the complete transcript and the baseline.
Similarly, each third alone was often sufficient to outperform the baseline, suggesting that the signal is redundantly present throughout the transcript (Appendix~\ref{app:thirds-kept-alone}).

\paragraph{Reasoning and tool interactions carry capability signal in some domains.}
\label{sec:results-source-channels}

We next ask whether capability-relevant signal is concentrated in specific channels of information: the model’s reasoning, its messages, or its tool interactions. To test this, we remove each channel in turn and compare the prediction error with that obtained from the complete transcript.

\begin{figure}[!htbp]
    \centering
    \includegraphics[width=.84\textwidth]{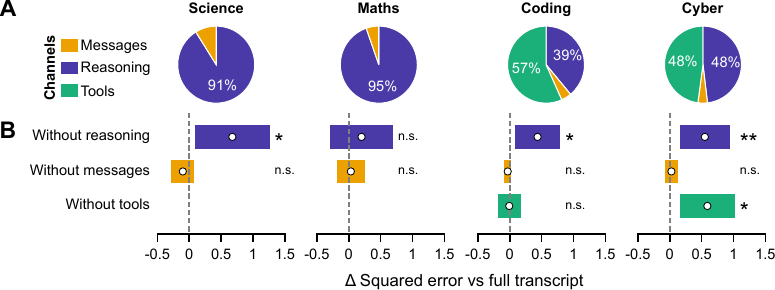}
    \caption{\textbf{Effect of removing transcript channels on prediction error.}
        \textbf{A.} Share of transcript in each channel per domain.
        \textbf{B.} Change in prediction error of pairwise judging when one channel is removed from every transcript, relative to the complete transcript (dashed line). Each point is the mean over a domain's target models and the box its 95\% paired $t$-interval.
        *: $p<.05$, **: $p<.01$, paired $t$-test of squared error between channel-removed and full transcripts.
        }
    \vspace{-.5em}
    \label{fig:channels}
\end{figure}

Reasoning and tool interactions carried capability-relevant signal in some domains, whereas messages did not. Removing reasoning increased prediction error in \code, \cyber, and \science (MSE changes: 0.18 to 0.62; 0.20 to 0.75; 0.27 to 0.94, all $p<.05$), but not significantly in \maths (0.31 to 0.51, $p=.40$). Removing tool interactions increased error in \cyber{} (0.20 to 0.79, $p=.01$) but had no effect in \code (0.18 to 0.17, $p=.93$). Removing messages did not significantly affect prediction error in any domain (Figure~\ref{fig:channels}B).

These effects are not explained by channel size alone. Tool interactions make up about half of both \code and \cyber transcripts, and reasoning over 90\% of both \maths and \science transcripts (Figure~\ref{fig:channels}A), yet removing them only degraded prediction in \cyber and \science, respectively.

\subsection{Generalisation of capability predictions}
\label{sec:results-generalization}

Our results so far show that saturated benchmark transcripts contain capability-predicting signals that pairwise judges can extract. We demonstrate this in a setting where protocols predict capabilities of newer target models one semester ahead. This prediction may involve several kinds of generalisation: newer models may be trained differently, reach higher capabilities, or solve tasks in different ways. In practice, protocols may also need to generalise beyond a single semester -- for example, by looking further ahead in time, or predicting capabilities in domains they were not calibrated on. We therefore test how prediction performance changes across time, model classes, capability levels, and domains.

\paragraph{Prediction reliability varies across domains and historical periods.}
We first test whether protocols' performances are stable across time. We repeat predictions retrospectively across earlier target periods, each time using models from the preceding period as the reference set. We back-test \science and \maths to early 2024; \code and \cyber begin in late 2025 because earlier models do not provide sufficient coverage of these agentic benchmarks.

\begin{figure}[!htbp]
    \centering
    \includegraphics[width=.8\linewidth]{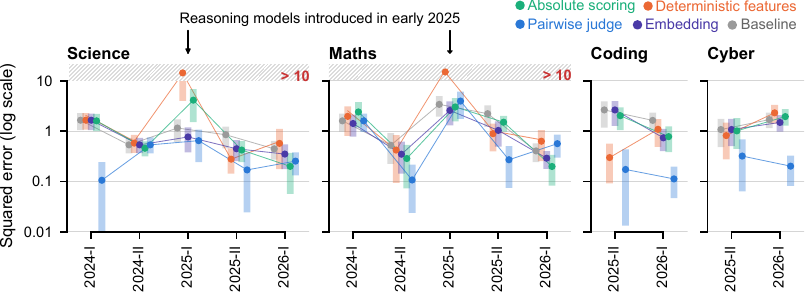}
    \caption{
        \textbf{Prediction errors vary substantially across historical target periods.}
        Prediction error for each protocol when calibrated on the preceding reference population and evaluated on the 8 models of each target period. Values above 10 are drawn in the hatched band.
    }
    \label{fig:time_generalisation}
    \vspace{-1em}
\end{figure}

Prediction error varies substantially across historical periods for all protocols (Figure~\ref{fig:time_generalisation}, Appendix~\ref{app:back-testing}). Pairwise judging performs consistently well in \code and \cyber, but is less reliable in \maths and \science, where it fails to improve on the baseline in four of ten evaluated periods (\maths: early 2024, early 2025, and early 2026; \science: late 2024). Other protocols show similar temporal instability. The most pronounced failures occur in early 2025, particularly for deterministic transcript features, where errors surge in both \science and \maths.

\paragraph{Pre-reasoning-era judges underestimate the capability of reasoning models.}
The early-2025 failures coincide with the introduction of reasoning models. This marks a deviation from reference models,
accompanied by a sharp increase in mathematical capabilities. We therefore use this period as a case study to assess how this influences prediction performance for pairwise judging.

\begin{SCfigure}[1][!h]
    \centering
    \includegraphics[width=.57\textwidth]{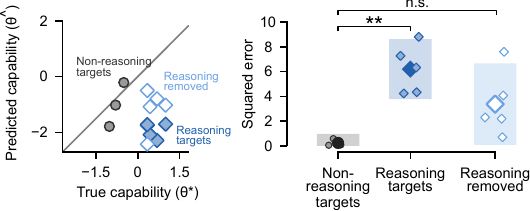}
    \caption{
        \textbf{Pairwise judges underestimate maths capabilities of early-2025 reasoning models.}
        \textbf{Left,} Predicted vs true capability for the three non-reasoning targets, five reasoning targets, and reasoning targets judged with reasoning withheld. \textit{Diagonal}: perfect prediction.
        \textbf{Right,} Prediction error per group (means $\pm$ 95\% $t$-intervals). **: $p<.01$, two-sample $t$-test.
    }
    \label{fig:reasoning_shift}
\end{SCfigure}

Prediction error is lower for the three non-reasoning targets than for the five reasoning targets (MSE 0.24 vs.\ 6.20; $p=.002$; Figure~\ref{fig:reasoning_shift}). This was because judges underestimated the capabilities of reasoning models, with predictions falling 2.1--3.0 $\theta$ units below their measured values. This failure is not fully explained by the additional reasoning text: removing the reasoning tokens from the transcripts only approximately halves the error to 3.37 ($p=.16$): still higher than for non-reasoning targets. These results suggest that the shift to reasoning models changes the relationship between transcript behaviour and capability in ways that the pairwise judging protocol is unable to capture.

\paragraph{Pairwise judging predictions generalise to more capable models.}
\label{sec:capability-gen}

\begin{figure}[!h]
    \centering
    \includegraphics[width=0.95\textwidth]{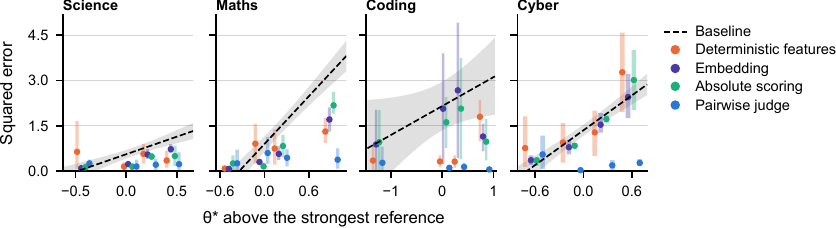}
    \caption{\textbf{Pairwise judging generalises across target model capabilities.} Prediction error as target capabilities exceed that of the strongest reference model. Targets are grouped into four equal-count bins. Error increases with capability distance for all protocols except pairwise judging.}
    \vspace{-.5em}
    \label{fig:capability_diff}
\end{figure}

Next, we test whether prediction quality degrades as target model capabilities exceed the reference models used for calibration.
For each target, we measure how far its capability exceeds that of the strongest reference model. Pairwise judging remains stable as this gap increases (Figure~\ref{fig:capability_diff}): the slope of squared error against capability gap is indistinguishable from zero in every domain (\science: $-0.02$, \maths: $0.10$, \code: $-0.06$, \cyber: $-0.12$; all $p \geq .45$). All other protocols show increasing error as targets move beyond the reference capability range.

\begin{wrapfigure}{R}{0.4\textwidth}
    \centering
    \includegraphics[width=\linewidth]{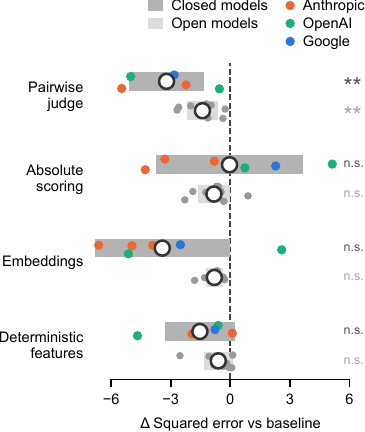}
    \caption{
    \textbf{Pairwise judge predictions generalise to closed-source targets.}
    Change in squared error from each target's baseline error (below zero: better than the baseline protocol) for six closed-source targets (coloured) and eight open-weight targets from the same semester.
    }
    \vspace{-1.5em}
    \label{fig:open-closed}
\end{wrapfigure}

\paragraph{Pairwise judge predictions transfer to closed-source models.}
\label{sec:closed_abl}
In practice, capability estimation is especially important for frontier models, which are often closed-source. Although pairwise judging generalises well to more capable open-weight models (Figure~\ref{fig:capability_diff}), closed-source frontier models introduce two additional shifts: they may differ in training and behaviour, and their reasoning traces are typically unavailable. The latter is particularly relevant, since removing reasoning has been shown to notably reduce capability signal (Section~\ref{sec:results-source-channels}). We therefore evaluate six closed-source targets on \code (Claude Opus 4.7, Opus 4.8, Fable 5, GPT-5, GPT-5.6 Sol, and Gemini 3.5 Flash), and compare capability predictions on these targets with those of open-weight targets from the most recent semester (early 2026).
We evaluate protocols with reasoning hidden for all models.

Performance transferred to closed-source targets to varying degrees (Figure~\ref{fig:open-closed}). Pairwise judging beat the baseline protocol on all six closed-source targets and all eight open-weight targets. No other protocol performed significantly better than the baseline for either model group.

\paragraph{Pairwise judge predictions are strongest in \code and \cyber.}
The analysis in Section~\ref{sec:results-signal} averages performance across domains, potentially obscuring domain-specific differences between protocols. We therefore examine protocol performance separately by domain (Table~\ref{tab:domain-mse}). Pairwise judging achieves the lowest prediction error amongst protocols in all four domains, but its improvement over the baseline is only significant in the agentic domains (\code: mean LLR 1.28, $p<.001$; \cyber: 0.94, $p=.001$). The remaining protocols show limited generalisation across domains. We report further analysis of domain dependence in Appendix~\ref{app:per-domain} and cross-domain transfer results in Appendix~\ref{app:domain-transfer}.

\begin{table}[!h]
    \centering
    \small
    \caption{
        \textbf{Prediction error per protocol and domain.}
        MSE per domain. Lowest prediction errors per domain are bolded. */**/***: $p<.05/.01/.001$, one-sample $t$-tests on LLR.
    }
    \label{tab:domain-mse}
    \begin{tabular}{lccccc}
        \toprule
        \textbf{Protocol} & \textbf{Pooled} & \textbf{\science} & \textbf{\maths} & \textbf{\code} & \textbf{\cyber} \\
        \midrule
        Baseline protocol & 1.37 & 0.65 & 1.32 & 2.14 & 1.39 \\
        Deterministic features & 0.86 & 0.43 & 0.76 & 0.70* & 1.57 \\
        Embedding & 1.01 & 0.40** & 0.66 & 1.69 & 1.29** \\
        Absolute scoring & 1.01 & 0.31 & 0.85 & 1.41 & 1.48 \\
        Pairwise judging & \textbf{0.26}** & \textbf{0.21} & \textbf{0.42} & \textbf{0.14}*** & \textbf{0.26}** \\
        \bottomrule
    \end{tabular}
    \vspace{-1em}
\end{table}

\section{Related Work}
\label{sec:related-work}

\textbf{Benchmark saturation and evaluation design.}
Previous works on benchmarks report the impact of score saturation and the downstream limit to discrimination amongst strong models \citep{bowman2021fix,vania2021comparing,akhtar2026saturation}.
Attempts to resolve this have typically involved introducing new challenging versions of existing benchmarks \citep{hendrycks2021math,suzgun2023challenging,rein2024gpqa,wang2024mmlupro,phan2026hle}. Other lines of work also explore the idea of dynamic evaluation, which instead uses adversarial data collection or some continuously updated test set to maintain difficulty over time \citep{kiela2021dynabench,white2025livebench,jain2024livecodebench}, but costs for running these methods are much higher than alternatives.

\textbf{Information from model responses.}
Prior works on transcript analysis extract capability-relevant signals through pairwise comparison and rubric-based scoring \citep{zheng2023judge,liu2023geval,kim2024prometheus2}, process-level reasoning assessment \citep{uesato2022solving,lightman2024lets}, and confidence-based prediction of correctness \citep{lin2022teaching,kadavath2022language,tian2023just}. These approaches point to model transcripts as a promising source of information. To our knowledge, no previous work has explored whether correctly solved transcripts from saturated benchmarks can signal capabilities.

\textbf{Capability estimation and prediction.}
Aggregate benchmark scores discard information about how models perform on individual items, which can lead to poorer predictions of model performance than per-item analysis \citep{schellaert_proposal_2024}.
A common approach for this is item response theory (IRT), used for latent ability estimation, item calibration, and test-set comparison \citep{lalor2016building,lalor2019learning,rodriguez2021evaluation,vania2021comparing,zhou2026lost}. By accounting for differences in item difficulty, IRT can be used to provide a more structured capability estimate than accuracy~\citep{baker1998item}.
Additional works on adaptive testing and representative item selection focus on a different direction of reducing evaluation cost by identifying informative subsets of tasks \citep{zhuang2025position,polo2024tinybenchmarks,fonseca2026irt}, although estimation reliability depends on the number and distribution of evaluated models \citep{jiang2026trust}. Finally, forecasting work predicts benchmark performance from signals on other tasks \citep{owen2024predictable,ruan2024observational}.

\section{Discussion}
\label{sec:discussion}

We explore whether benchmarks can remain useful for evaluating model capabilities, even after final scores are saturated.
We show that transcripts from saturated benchmarks retain capability-predicting information.
Amongst the protocols we assessed, pairwise judging was able to extract this information most reliably, outperforming predictions from saturated benchmark scores in predicting performance on more difficult held-out tasks.
More broadly, these results suggest that it is possible to extend the usefulness of saturated benchmarks for tracking model capabilities: a critical capacity for informing safety and governance decisions, when constructing harder evaluations cannot keep pace with model progress.

Demonstrating that capability information is present in transcripts is of course a first step. A useful capability-estimation protocol must also extract this information reliably when the relationship between transcript behaviour and capability changes; our current protocols do not yet meet this requirement. Prediction performance varies substantially across time, and the transition to reasoning models in early 2025 provides a clear failure case in which judging-based protocols systematically underestimate the capabilities of the new model class. Future work should focus not only on extracting more signal from transcripts, but on designing protocols that are robust to distribution shifts.

\section*{Acknowledgments}
This work was funded by Coefficient Giving through the LASR Labs fellowship. We thank Tamkeen Nawab, Brandon Riggs, Alexander Barry, and all LASR programme participants for their advice and feedback; James Mann and other Generality Labs members for helpful discussions.

\bibliographystyle{plainnat}
\bibliography{custom}

\appendix

\renewcommand{\sectionautorefname}{Appendix}
\renewcommand{\subsectionautorefname}{Appendix}

\clearpage
\section*{Appendix Contents}
\startcontents[appendix]
\printcontents[appendix]{l}{1}{\setcounter{tocdepth}{2}}
\clearpage
\newpage
\section{Detailed Methodology}
\label{app:methodology_full}

\subsection{Quantification of protocol predictions}
\label{app:prediction-stats}

\begin{figure}[!h]
      \centering
      \includegraphics[width=0.7\linewidth]{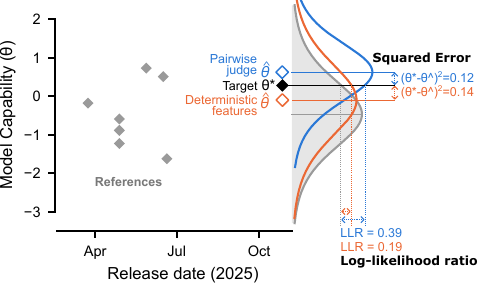}
      \caption{\textbf{Quantification of squared error and log-likelihood ratio for an example target model.}
      Prediction of one target model's capability (\textit{black}: measured $\theta_m^*$) from reference models in the preceding semester (\textit{grey diamonds}), plotted against release date. For each protocol, predictive uncertainty is represented by a Student-$t$ distribution with $n-1$ degrees of freedom, centred on the protocol's prediction and scaled using leave-one-model-out prediction errors on the reference models (\textit{grey}: baseline protocol; \textit{blue}: pairwise judging; \textit{orange}: deterministic features). Pairwise judging and deterministic features miss $\theta_m^*$ by similar amounts in opposite directions (squared error 0.12 and 0.14, respectively), but the narrower distribution for pairwise judging assigns greater density to $\theta_m^*$, resulting in a larger log-likelihood ratio relative to the baseline protocol (0.39 versus 0.19).}
\label{fig:prediction_example}
\end{figure}

For each prediction protocol $\mathcal{P}$ and target model $m$, we first measure prediction accuracy using squared error:
\begin{equation}
    \epsilon_{m,\mathcal{P}}
    =
    \left(
    \hat{\theta}_{m,\mathcal{P}}-\theta_m^*
    \right)^2.
\end{equation}

We additionally estimate how stable each protocol's predictions are given the finite reference set. For each of the $n$ reference models $r$ (7 or 8 per domain and semester), we refit protocol $\mathcal{P}$ on the remaining $n-1$ references and predict the held-out reference capability $\theta_r^*$. We summarise these leave-one-model-out errors using
\begin{equation}
    s_{\mathcal{P}}
    =
    \sqrt{
    \frac{1}{n}
    \sum_{r\in\mathcal{M}_{\mathrm{ref}}}
    \left(
    \hat{\theta}^{(-r)}_{r,\mathcal{P}}-\theta_r^*
    \right)^2
    }.
\end{equation}

Thus, $s_{\mathcal{P}}$ estimates the protocol's out-of-sample prediction error on models drawn from the reference population.

We use this quantity to model predictive uncertainty for each target. Specifically, we define a Student-$t$ distribution with $n-1$ degrees of freedom, centred on the target prediction $\hat{\theta}_{m,\mathcal{P}}$ and scaled by $s_{\mathcal{P}}$:
\begin{equation}
  p_{\mathcal{P}}(\theta)
  =
  \frac{1}{s_{\mathcal{P}}}
  t_{n-1}
  \left(
  \frac{\theta-\hat{\theta}_{m,\mathcal{P}}}{s_{\mathcal{P}}}
  \right).
\end{equation}

We use a Student-$t$ rather than a normal distribution because the prediction scale is estimated from only 7 or 8 reference models per domain and semester. We apply the same construction to the baseline protocol.

For each target, we then compute the log-likelihood ratio (LLR) between the protocol and baseline predictive distributions:
\begin{equation}
  \mathrm{LLR}_{m,\mathcal{P}}
  =
  \log p_{\mathcal{P}}(\theta_m^*)
  -
  \log p_{\mathrm{baseline}}(\theta_m^*).
\end{equation}

A positive LLR indicates that the protocol assigns greater predictive density to the target's measured capability than the baseline (Figure~\ref{fig:prediction_example}). Unlike squared error, the LLR also accounts for the protocol's estimated prediction uncertainty: two protocols with similar squared error can receive different LLRs if one is substantially more or less stable across the reference models.

\subsection{Item splitting and capability measurement}
\label{app:irt}

For each domain, we split items and estimate model capabilities as follows:

\begin{enumerate}
    \item \textbf{Responses.} We collect every model's scored response to every item in the domain's benchmarks.

    \item \textbf{Global IRT fit.} We fit a two-parameter logistic IRT model with a fixed guessing floor to all model responses. Each item has a discrimination parameter $a_i$, a difficulty parameter $b_i$, and a fixed guessing floor $c_i$, set to $1/k$ for a multiple-choice item with $k$ options and to 0 for a free-response item. Each model receives a global capability estimate.

    \item \textbf{Reference and target sets.} For each target semester, the reference set consists of models from the preceding semester, while the target set consists of models from the target semester.

    \item \textbf{Saturation boundary.} The three reference models with the highest global capability define the split. An item is \emph{saturated} if all three answer it correctly, and \emph{unsaturated} if at least one answers it incorrectly. We ablate this in Appendix~\ref{app:topk}.

    \item \textbf{Judged items.} From the saturated set, we draw a fixed random sample of up to 25 items from each quartile of item difficulty $b_i$, for a maximum of 100 items. For section and channel removal (Section~\ref{sec:results-source}), we used a 40-item subset (10 per quartile). The same candidate items are used for every model and protocol; a model contributes a transcript for an item only when it answers that item correctly.

    \item \textbf{Measured capability.} For every reference and target model, we estimate $\theta^*$ using only the unsaturated items while holding $a_i$, $b_i$, and $c_i$ fixed at their global values. This places all semester-specific capability estimates within a domain on a common scale.
\end{enumerate}

Protocols observe only correctly solved transcripts from the sampled saturated items and are evaluated on how well they predict $\theta^*$ measured on the unsaturated items.

\clearpage

\section{Additional Experimental Configuration}
\label{app:experimental-config}

This appendix provides additional details on the models and benchmark data used in our retrospective evaluations.

\subsection{Evaluated Models}
\label{app:evaluated-models}

We organise our evaluation framework into six-month periods (semesters). From each semester, we select eight models: these serve as both the target models (for protocols calibrated on the preceding semester) and the reference models for target models in the next semester. For our main analyses, we restricted our model selection to only those that show unencrypted reasoning traces (but see Section~\ref{sec:closed_abl} for additional experiments with closed-source models with hidden reasoning). We include closed-source models in 2023 and 2024, as open-weight reasoning models with visible traces were not yet available in these periods.

Table~\ref{tab:evaluated-models} lists the models included in each semester. Within each period, we select models spanning a range of capability levels rather than only the strongest available models to penalise strategies that always estimate high capabilities.
The capability is estimated based on the accuracy on a set of collected benchmarks and transcripts; the full collected data are described in Appendix~\ref{app:benchmark-data}.

Within each period and domain, the most capable model (derived from IRT across all items in a domain; see Appendix~\ref{app:irt}) is shown in bold and the second and third are underlined; closed-source models are shown in italics. For models with explicit reasoning-effort controls, we use the provider default where available and ``medium'' otherwise. A dash indicates that the model is not evaluated in that domain.

\begin{table*}[!h]
  \centering
  \caption{
    \textbf{Models used in the retrospective evaluation.}
    Models are grouped by evaluation period. Entries give the estimated global capability $\theta$ $\pm$ the half-width of its 95\% credible interval. Within each period and domain, the highest $\theta$ is in bold and the second and third are underlined, with shading of matching strength. Closed-source models are shown in italics. A dash indicates that the model is not evaluated in that domain.
  }
  \label{tab:evaluated-models}
  \small
  \definecolor{pw}{HTML}{2a78d6}
  \setlength{\tabcolsep}{0.3em}
  \begin{tabular}{llcccc}
    \toprule
    \textbf{Period} & \textbf{Model} & \science & \maths & \code & \cyber \\
    \midrule

    \multirow{8}{*}{2023}
    & \textit{gpt-4-0613} & \cellcolor{pw!35}{\boldmath$-0.91$ {\scriptsize$\pm0.10$}} & \cellcolor{pw!35}{\boldmath$-1.85$ {\scriptsize$\pm0.21$}} & -- & -- \\
    & mixtral-8x7b-instruct-v0.1 & \cellcolor{pw!12}\underline{$-1.27$ {\scriptsize$\pm0.12$}} & \cellcolor{pw!12}\underline{$-2.75$ {\scriptsize$\pm0.28$}} & -- & -- \\
    & \textit{gpt-3.5-turbo-0125} & $-1.68$ {\scriptsize$\pm0.13$} & \cellcolor{pw!12}\underline{$-2.52$ {\scriptsize$\pm0.25$}} & -- & -- \\
    & nous-hermes-2-yi-34b & \cellcolor{pw!12}\underline{$-1.64$ {\scriptsize$\pm0.17$}} & $-3.07$ {\scriptsize$\pm0.31$} & -- & -- \\
    & qwen-72b-chat & $-2.03$ {\scriptsize$\pm0.18$} & $-2.90$ {\scriptsize$\pm0.29$} & -- & -- \\
    & llama-2-70b-chat-hf & $-1.85$ {\scriptsize$\pm0.13$} & $-3.18$ {\scriptsize$\pm0.32$} & -- & -- \\
    & llama-2-13b-chat-hf & $-2.16$ {\scriptsize$\pm0.17$} & $-4.23$ {\scriptsize$\pm0.63$} & -- & -- \\
    & mpt-30b & $-2.60$ {\scriptsize$\pm0.18$} & $-4.52$ {\scriptsize$\pm0.67$} & -- & -- \\
    \midrule

    \multirow{8}{*}{2024-I}
    & \textit{gpt-4o-2024-05-13} & \cellcolor{pw!35}{\boldmath$-0.16$ {\scriptsize$\pm0.09$}} & \cellcolor{pw!35}{\boldmath$-1.40$ {\scriptsize$\pm0.20$}} & -- & -- \\
    & \textit{gpt-4-turbo-2024-04-09} & \cellcolor{pw!12}\underline{$-0.31$ {\scriptsize$\pm0.10$}} & \cellcolor{pw!12}\underline{$-1.41$ {\scriptsize$\pm0.20$}} & -- & -- \\
    & qwen2-72b-instruct & $-0.63$ {\scriptsize$\pm0.10$} & \cellcolor{pw!12}\underline{$-1.53$ {\scriptsize$\pm0.21$}} & -- & -- \\
    & wizardlm-2-8x22b & \cellcolor{pw!12}\underline{$-0.57$ {\scriptsize$\pm0.17$}} & $-1.78$ {\scriptsize$\pm0.21$} & -- & -- \\
    & meta-llama-3-70b-instruct & $-0.62$ {\scriptsize$\pm0.10$} & $-1.91$ {\scriptsize$\pm0.22$} & -- & -- \\
    & gemma-2-27b-it & $-0.89$ {\scriptsize$\pm0.10$} & $-1.96$ {\scriptsize$\pm0.23$} & -- & -- \\
    & qwen1.5-72b-chat & $-1.24$ {\scriptsize$\pm0.12$} & $-2.32$ {\scriptsize$\pm0.23$} & -- & -- \\
    & phi-3-medium-128k-instruct & $-1.34$ {\scriptsize$\pm0.12$} & $-2.50$ {\scriptsize$\pm0.25$} & -- & -- \\
    \midrule

    \multirow{8}{*}{2024-II}
    & deepseek-v3 & \cellcolor{pw!35}{\boldmath$0.09$ {\scriptsize$\pm0.09$}} & \cellcolor{pw!35}{\boldmath$-0.73$ {\scriptsize$\pm0.18$}} & -- & -- \\
    & phi-4 & \cellcolor{pw!12}\underline{$-0.02$ {\scriptsize$\pm0.09$}} & \cellcolor{pw!12}\underline{$-1.16$ {\scriptsize$\pm0.18$}} & -- & -- \\
    & llama-3.3-70b-instruct & $-0.32$ {\scriptsize$\pm0.09$} & \cellcolor{pw!12}\underline{$-1.21$ {\scriptsize$\pm0.20$}} & -- & -- \\
    & qwen2.5-72b-instruct & \cellcolor{pw!12}\underline{$-0.22$ {\scriptsize$\pm0.09$}} & $-1.33$ {\scriptsize$\pm0.19$} & -- & -- \\
    & qwen2.5-32b-instruct & $-0.35$ {\scriptsize$\pm0.09$} & $-1.22$ {\scriptsize$\pm0.21$} & -- & -- \\
    & hermes-3-llama-3.1-405b & $-0.27$ {\scriptsize$\pm0.15$} & $-1.65$ {\scriptsize$\pm0.21$} & -- & -- \\
    & llama-3.1-8b-instruct & $-1.37$ {\scriptsize$\pm0.12$} & $-2.00$ {\scriptsize$\pm0.24$} & -- & -- \\
    & open-mistral-nemo-2407 & $-1.35$ {\scriptsize$\pm0.13$} & $-2.54$ {\scriptsize$\pm0.26$} & -- & -- \\
    \midrule

    \multirow{8}{*}{2025-I}
    & minimax-m1 & \cellcolor{pw!35}{\boldmath$0.98$ {\scriptsize$\pm0.14$}} & \cellcolor{pw!12}\underline{$0.46$ {\scriptsize$\pm0.14$}} & \cellcolor{pw!35}{\boldmath$0.87$ {\scriptsize$\pm0.25$}} & \cellcolor{pw!12}\underline{$0.70$ {\scriptsize$\pm0.40$}} \\
    & deepseek-r1-0528 & \cellcolor{pw!12}\underline{$0.92$ {\scriptsize$\pm0.10$}} & \cellcolor{pw!35}{\boldmath$0.76$ {\scriptsize$\pm0.14$}} & \cellcolor{pw!12}\underline{$-0.43$ {\scriptsize$\pm0.23$}} & \cellcolor{pw!35}{\boldmath$0.90$ {\scriptsize$\pm0.33$}} \\
    & qwen3-235b-a22b & \cellcolor{pw!12}\underline{$0.64$ {\scriptsize$\pm0.09$}} & \cellcolor{pw!12}\underline{$0.44$ {\scriptsize$\pm0.15$}} & \cellcolor{pw!12}\underline{$-1.15$ {\scriptsize$\pm0.29$}} & \cellcolor{pw!12}\underline{$0.10$ {\scriptsize$\pm0.36$}} \\
    & qwen3-32b & $0.35$ {\scriptsize$\pm0.11$} & $0.11$ {\scriptsize$\pm0.12$} & $-1.28$ {\scriptsize$\pm0.27$} & $0.04$ {\scriptsize$\pm0.28$} \\
    & deepseek-v3-0324 & $0.48$ {\scriptsize$\pm0.10$} & $-0.54$ {\scriptsize$\pm0.15$} & $-1.35$ {\scriptsize$\pm0.27$} & $0.03$ {\scriptsize$\pm0.36$} \\
    & qwen3-30b-a3b & $0.37$ {\scriptsize$\pm0.12$} & $0.08$ {\scriptsize$\pm0.14$} & $-1.31$ {\scriptsize$\pm0.27$} & $-0.75$ {\scriptsize$\pm0.30$} \\
    & gemma-3-27b-it & $-0.34$ {\scriptsize$\pm0.15$} & $-1.11$ {\scriptsize$\pm0.17$} & -- & -- \\
    & mistral-small-3.2-24b-instruct & $-0.18$ {\scriptsize$\pm0.13$} & $-0.88$ {\scriptsize$\pm0.16$} & $-1.72$ {\scriptsize$\pm0.29$} & $-1.19$ {\scriptsize$\pm0.30$} \\
    \midrule

    \multirow{8}{*}{2025-II}
    & kimi-k2-thinking & \cellcolor{pw!35}{\boldmath$1.55$ {\scriptsize$\pm0.14$}} & \cellcolor{pw!35}{\boldmath$1.64$ {\scriptsize$\pm0.16$}} & \cellcolor{pw!12}\underline{$1.07$ {\scriptsize$\pm0.25$}} & \cellcolor{pw!12}\underline{$1.01$ {\scriptsize$\pm0.39$}} \\
    & deepseek-v3.2 & $1.13$ {\scriptsize$\pm0.12$} & $1.28$ {\scriptsize$\pm0.16$} & \cellcolor{pw!12}\underline{$0.95$ {\scriptsize$\pm0.26$}} & \cellcolor{pw!35}{\boldmath$1.30$ {\scriptsize$\pm0.36$}} \\
    & glm-4.7 & \cellcolor{pw!12}\underline{$1.31$ {\scriptsize$\pm0.15$}} & \cellcolor{pw!12}\underline{$1.61$ {\scriptsize$\pm0.17$}} & $0.77$ {\scriptsize$\pm0.24$} & \cellcolor{pw!12}\underline{$0.74$ {\scriptsize$\pm0.39$}} \\
    & glm-4.5 & $1.23$ {\scriptsize$\pm0.14$} & \cellcolor{pw!12}\underline{$1.33$ {\scriptsize$\pm0.15$}} & $0.53$ {\scriptsize$\pm0.27$} & $0.67$ {\scriptsize$\pm0.36$} \\
    & minimax-m2 & \cellcolor{pw!12}\underline{$1.24$ {\scriptsize$\pm0.14$}} & $0.78$ {\scriptsize$\pm0.15$} & \cellcolor{pw!35}{\boldmath$1.12$ {\scriptsize$\pm0.25$}} & $0.48$ {\scriptsize$\pm0.38$} \\
    & qwen3-235b-a22b-thinking-2507 & $0.98$ {\scriptsize$\pm0.13$} & $1.05$ {\scriptsize$\pm0.16$} & $-0.80$ {\scriptsize$\pm0.26$} & $0.33$ {\scriptsize$\pm0.40$} \\
    & gpt-oss-120b & $0.65$ {\scriptsize$\pm0.13$} & $0.32$ {\scriptsize$\pm0.14$} & $-0.30$ {\scriptsize$\pm0.26$} & $0.22$ {\scriptsize$\pm0.35$} \\
    & gpt-oss-20b & $0.10$ {\scriptsize$\pm0.14$} & $0.16$ {\scriptsize$\pm0.14$} & $-0.90$ {\scriptsize$\pm0.26$} & $0.41$ {\scriptsize$\pm0.35$} \\
    \midrule

    \multirow{8}{*}{2026-I}
    & glm-5.1 & $1.50$ {\scriptsize$\pm0.16$} & $1.14$ {\scriptsize$\pm0.16$} & \cellcolor{pw!35}{\boldmath$1.91$ {\scriptsize$\pm0.26$}} & \cellcolor{pw!35}{\boldmath$2.12$ {\scriptsize$\pm0.42$}} \\
    & glm-5.2 & \cellcolor{pw!12}\underline{$1.68$ {\scriptsize$\pm0.13$}} & $1.43$ {\scriptsize$\pm0.16$} & \cellcolor{pw!12}\underline{$1.70$ {\scriptsize$\pm0.29$}} & \cellcolor{pw!12}\underline{$1.73$ {\scriptsize$\pm0.37$}} \\
    & kimi-k2.6 & \cellcolor{pw!35}{\boldmath$1.76$ {\scriptsize$\pm0.13$}} & \cellcolor{pw!35}{\boldmath$1.68$ {\scriptsize$\pm0.17$}} & \cellcolor{pw!12}\underline{$1.46$ {\scriptsize$\pm0.26$}} & $1.49$ {\scriptsize$\pm0.37$} \\
    & qwen3.5-397b-a17b & $1.48$ {\scriptsize$\pm0.15$} & \cellcolor{pw!12}\underline{$1.58$ {\scriptsize$\pm0.16$}} & $1.27$ {\scriptsize$\pm0.27$} & $1.19$ {\scriptsize$\pm0.40$} \\
    & kimi-k2.5 & \cellcolor{pw!12}\underline{$1.54$ {\scriptsize$\pm0.12$}} & $1.54$ {\scriptsize$\pm0.17$} & $1.25$ {\scriptsize$\pm0.27$} & $1.11$ {\scriptsize$\pm0.40$} \\
    & nvidia-nemotron-3-ultra-550b-a55b & $1.29$ {\scriptsize$\pm0.11$} & \cellcolor{pw!12}\underline{$1.55$ {\scriptsize$\pm0.15$}} & $0.86$ {\scriptsize$\pm0.23$} & $1.47$ {\scriptsize$\pm0.35$} \\
    & qwen3.6-27b & $1.22$ {\scriptsize$\pm0.14$} & $1.03$ {\scriptsize$\pm0.16$} & $1.13$ {\scriptsize$\pm0.25$} & \cellcolor{pw!12}\underline{$1.60$ {\scriptsize$\pm0.40$}} \\
    & gemma-4-31b-it & $1.33$ {\scriptsize$\pm0.12$} & $1.13$ {\scriptsize$\pm0.15$} & $1.03$ {\scriptsize$\pm0.26$} & $1.43$ {\scriptsize$\pm0.37$} \\

    \bottomrule
  \end{tabular}
\end{table*}

\clearpage

\subsection{Collected Benchmark Data}
\label{app:benchmark-data}

Our main evaluation retains benchmark items across four domains.

\paragraph{Science.}
We use all 198 questions in GPQA-Diamond.
For ARC-Challenge, we select a fixed random subset of 200 questions from the 1,172-question \texttt{test} split, using seed 0.
For HLE, only the 575 text-only questions in the HLE-Verified gold subset are used for analysis.
The science domain retains 210 questions tagged Physics, Chemistry, Biology/Medicine, Computer Science/AI, or Engineering.

\paragraph{Mathematics.}
We select 150 questions from the \texttt{math} category of the MMLU-Pro \texttt{test} split, using a fixed random sample with seed 42.
AIME 2024, AIME 2025, and AIME 2026 each contribute all 30 problems in their respective datasets.
We also use all 12 problems from APEX 2025 and all 47 problems from APEX Shortlist, each provided in a \texttt{train} split.
HLE-Gold contributes the 263 text-only gold questions tagged \texttt{Math}.

\paragraph{Coding.}
For AgentBench OS, we use the revised \texttt{test} split from Inspect's AgentBench OS task version 4-A.
It contains 119 tasks after six ill-posed tasks were removed from the original 125-task split.
For SWE-Bench, we use all 50 tasks in the \texttt{test} split of SWE-bench Verified Mini, distributed as \texttt{MariusHobbhahn/swe-bench-verified-mini}.
For Terminal-Bench, we use all 89 tasks from version 2.1, pinned to revision 6.

\paragraph{Cybersecurity.}
We use the hard variant of CyBench with 39 challenges; the Inspect implementation omits the \texttt{motp} challenge for licensing reasons.
For InterCode, we use Inspect's GDM InterCode-CTF implementation with 78 challenges.
This subset excludes 22 internet-dependent challenges from the original 100-task CTF collection.

\begin{figure}
    \centering
    \includegraphics[width=0.35\linewidth]{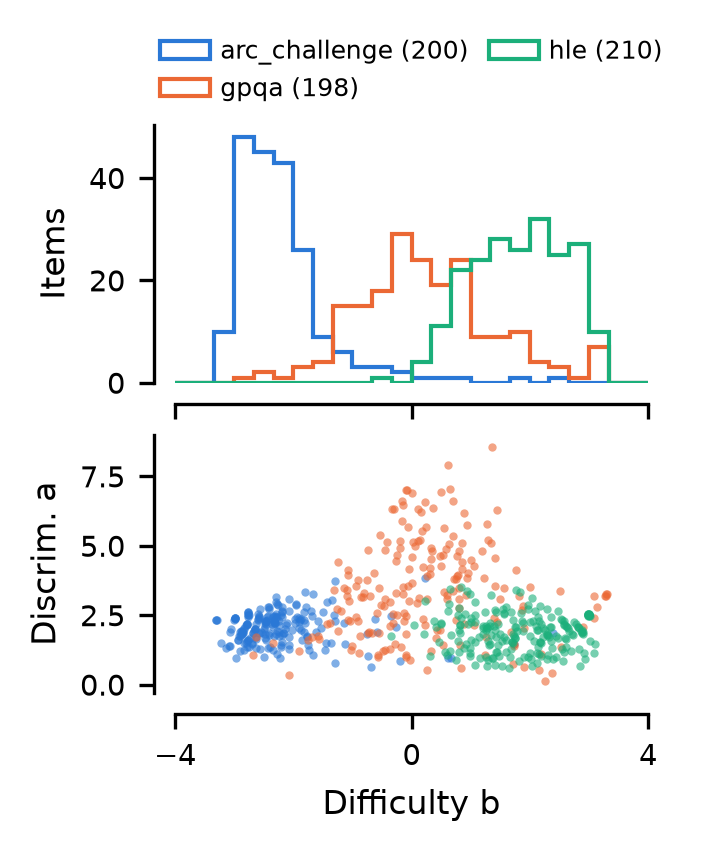}\hspace{2em}
    \includegraphics[width=0.4\linewidth]{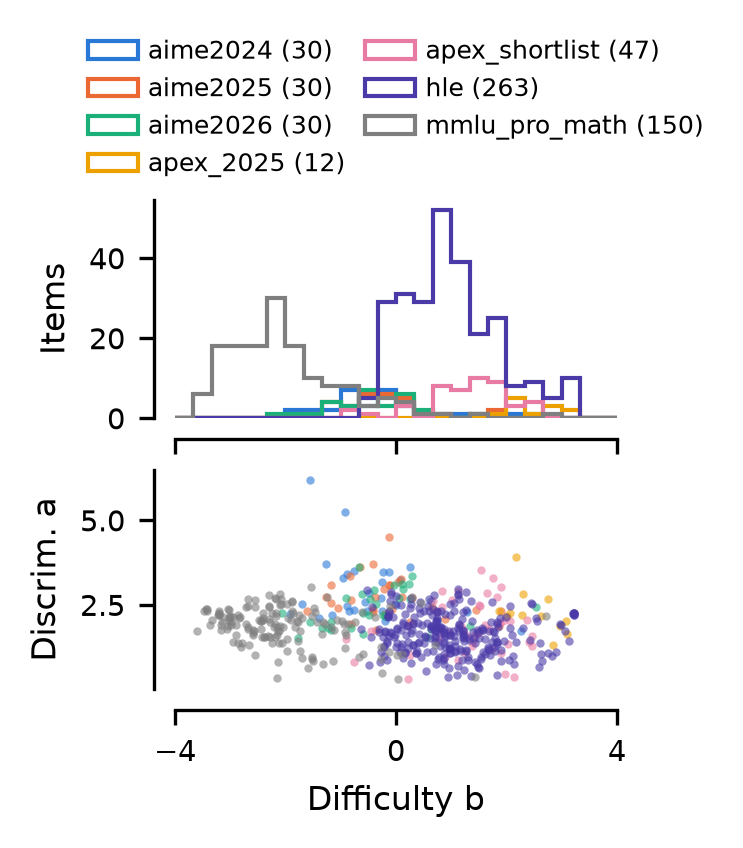}\par\medskip
    \includegraphics[width=0.42\linewidth]{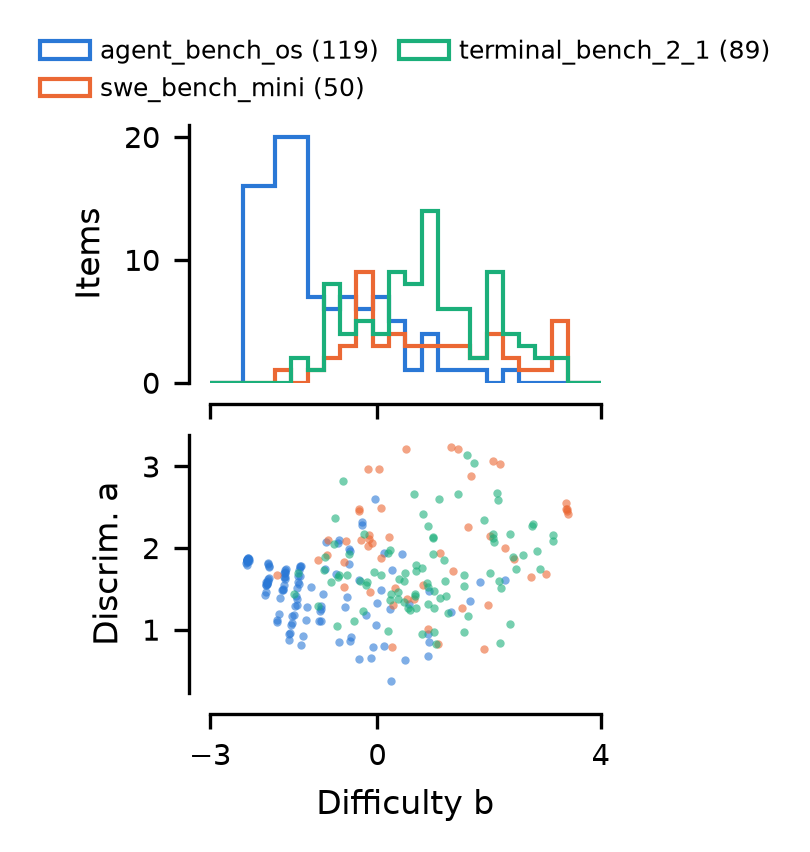}\hspace{2em}
    \includegraphics[width=0.35\linewidth]{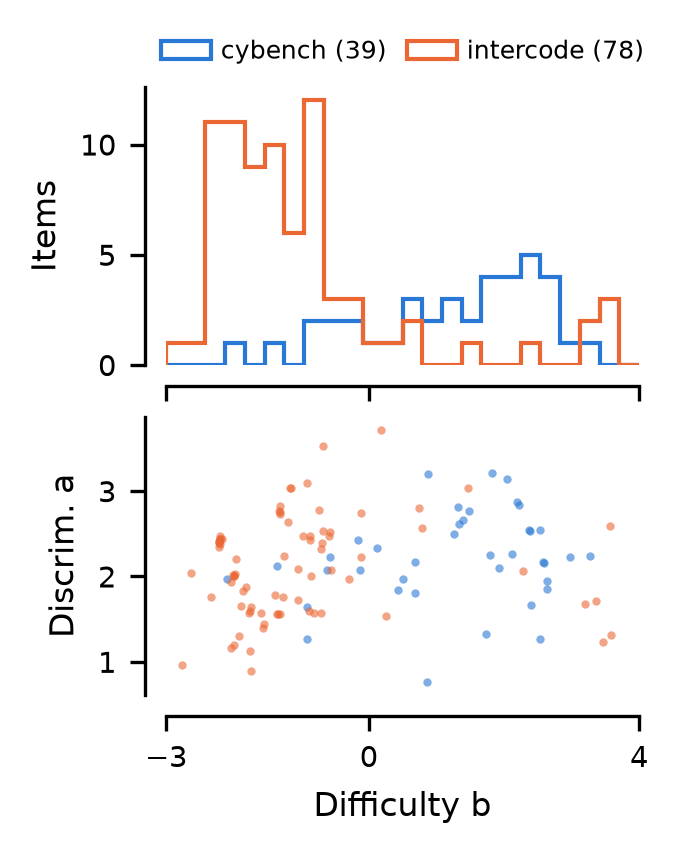}
    \caption{
    \textbf{Item difficulty parameters from the global IRT fit.}
    \science and \maths are shown in the top row; \code and \cyber in the bottom row. For each domain, the upper panel shows the distribution of item difficulty $b_i$ by benchmark, with the number of items shown in brackets. The lower panels show discrimination $a_i$ against difficulty $b_i$.
    }
    \label{fig:difficulty}
\end{figure}

\begin{figure}
    \centering
    \includegraphics[width=\linewidth]{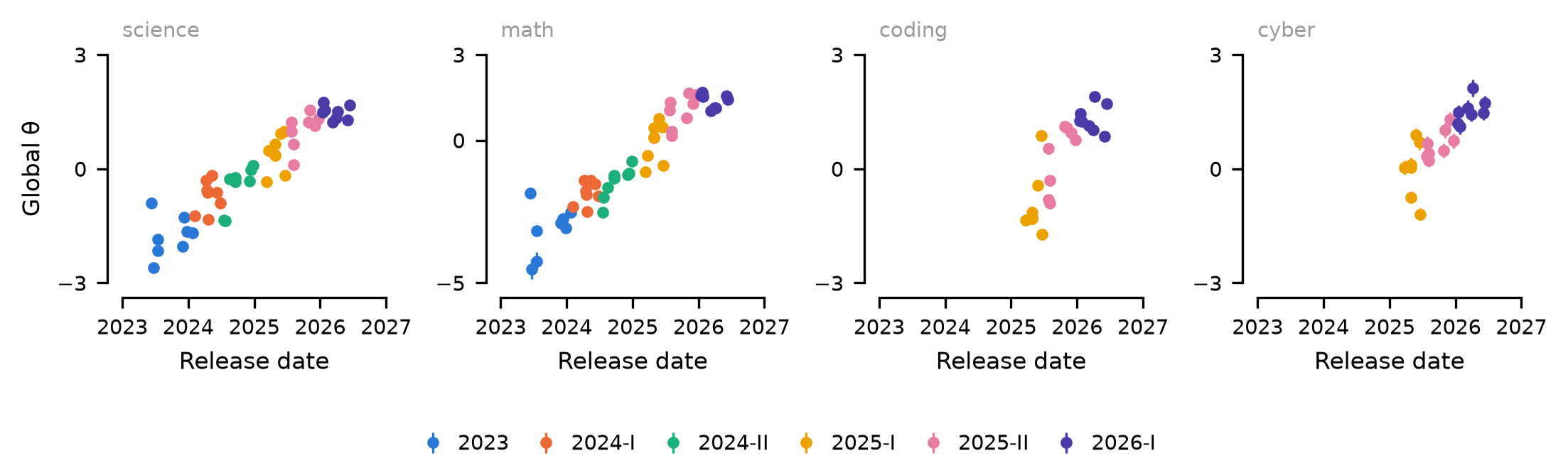}
    \caption{
    \textbf{Model capabilities by release date.}
    Estimated model capabilities in each domain against release date, based on the global IRT fit (see Appendix~\ref{app:irt}), coloured by evaluation period. Error bars show posterior standard deviation.
    }
    \label{fig:theta_release}
\end{figure}

\begin{figure}[t]
    \centering
    \includegraphics[width=\linewidth]{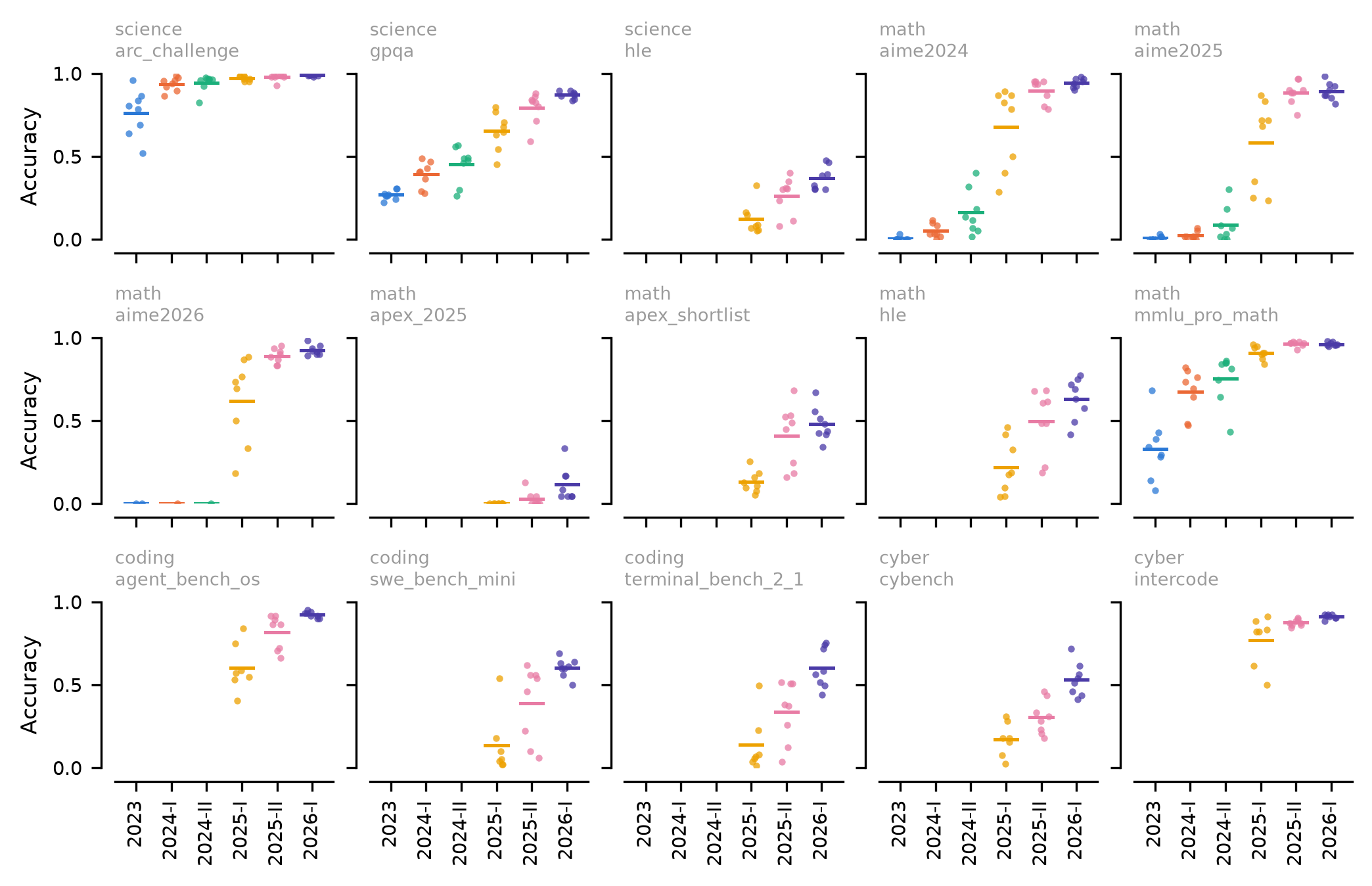}
    \caption{
    \textbf{Benchmark accuracy by evaluation period.}
    Points show individual models and lines show period means.
    }
    \label{fig:scores_vs_release_vs_domain}
\end{figure}

Figures~\ref{fig:accuracy}--\ref{fig:scores_vs_release_vs_domain} summarise the collected benchmark data. Accuracy generally increases across evaluation periods, causing the easier benchmarks to approach saturation for later models (Figure~\ref{fig:scores_vs_release_vs_domain}). The global IRT fit places benchmarks within each domain across distinct ranges of item difficulty (Figure~\ref{fig:difficulty}), while estimated global capability increases with model release date across all four domains (Figure~\ref{fig:theta_release}).

\begin{figure}[!h]
    \centering
    \caption{
    \textbf{Benchmark accuracy by model and benchmark.}
    Models are coloured by evaluation period. \science is shown here; the remaining domains continue on the following pages.
    \includegraphics[height=0.8\textheight]{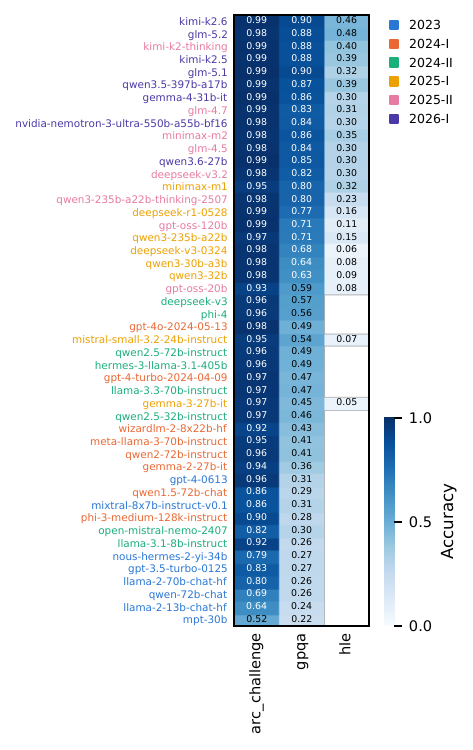}
    }
    \label{fig:accuracy}
\end{figure}

\begin{figure}[p]
    \ContinuedFloat
    \centering
    \includegraphics[height=0.7\textheight]{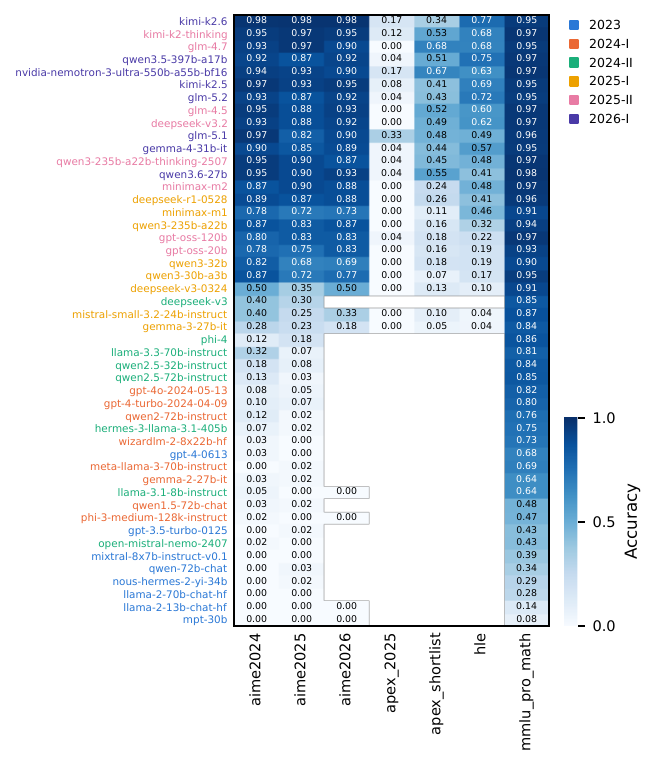}
    \caption{(Continued.) \maths.}
\end{figure}

\begin{figure}[p]
    \vspace{-2em}
    \ContinuedFloat
    \centering
    \includegraphics[height=0.47\textheight]{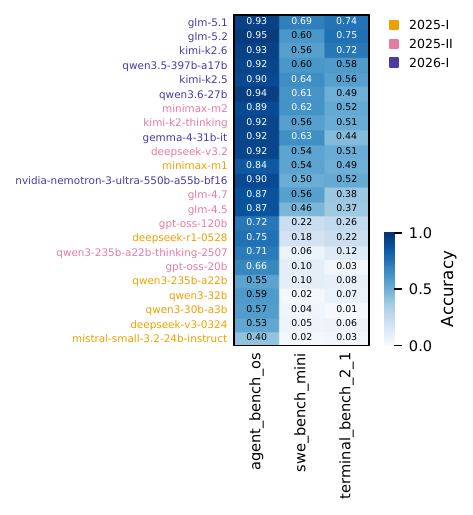}
    \vspace{-6em}
\end{figure}

\begin{figure}[p]
    \ContinuedFloat
    \centering
    \includegraphics[height=0.48\textheight]{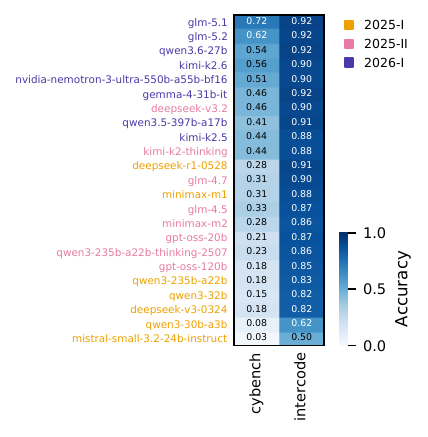}
    \caption{(Continued.) \code (top) and \cyber (bottom).}
\end{figure}

\clearpage

\section{Protocol Implementation}
\label{app:protocols}

This section describes the implementation of each prediction protocol. All protocols use the shared inputs and calibration data described below.

\subsection{Shared Inputs}
\label{app:protocols:shared}

\paragraph{Items.}
For each prediction task, we sample up to 25 saturated items from each quartile of global item difficulty $b_i$ using a fixed random seed, for a maximum of 100 items. Sampling is performed once at the item level so that all models and protocols use the same candidate items. For the position and channel analyses (Section~\ref{sec:results-source}; Appendices~\ref{app:thirds-kept-alone}, \ref{app:thirds-other-protocols}, and~\ref{app:channels-other-protocols}) and the reference-period sweep (Appendix~\ref{app:static_pairwise}), we use a fixed subset of 10 items per difficulty quartile. Protocols never observe unsaturated items or whether a model failed an item.

\paragraph{Transcripts.}
For each model--item pair, we use the first attempt and retain the transcript only if the model answers the item correctly. Transcripts contain the message history excluding the system message and initial user prompt; subsequent user messages in agentic tasks are retained. Assistant messages, reasoning traces, tool calls, and tool outputs are marked by role. Unless otherwise stated, visible reasoning is included. For multiple-choice tasks, each transcript is paired with the answer-option ordering shown to that model.

\paragraph{Calibration data.}
Each reference model $r \in \mathcal{M}_{\mathrm{ref}}$ is associated with its measured capability $\theta_r^*$ on the unsaturated items. Every protocol is calibrated using only the reference models for that prediction task, ensuring that all protocols and the baseline use the same labelled models. Target capabilities $\theta_m^*$ are used only for evaluation.

\paragraph{Baseline.}
The baseline assigns every target model the mean measured capability of the reference models:
\[
\hat{\theta}_m
=
\bar{\theta}_{\mathrm{ref}}
=
\frac{1}{|\mathcal{M}_{\mathrm{ref}}|}
\sum_{r \in \mathcal{M}_{\mathrm{ref}}}
\theta_r^*.
\]
Because protocols observe only correctly solved saturated items, benchmark accuracy provides no information for distinguishing target models in this setting.

\subsection{Pairwise Judging}
\label{app:protocols:pairwise}

\paragraph{Comparisons.}
For each benchmark, we compare every reference--target pair and every pair of reference models. Reference--reference comparisons calibrate the relationship between judge preferences and capability differences, while reference--target comparisons locate each target relative to the calibrated references. Target--target comparisons are not used. A pair is evaluated on an item only when both models answer it correctly and produce a non-trivial response. Each comparison is presented in both transcript orders to reduce sensitivity to position bias.

\paragraph{Judge prompt.}
The judge is an LLM sampled at temperature $0$. It receives the task followed by two correct transcripts and is asked which one was produced by the more capable model. For multiple-choice tasks, the answer-option ordering shown to each model is also provided. The prompts are reproduced verbatim; angle brackets mark filled-in fields, and the options blocks appear only for multiple-choice tasks.

\begin{promptbox}[colframe=pairwisecol, colbacktitle=pairwisecol!10]{Pairwise judge: system prompt}
You are an expert LLM evaluator, tasked with judging the responses of two AI models to the same task.
You will be given the question and the responses from two models, labeled Transcript 1 and Transcript 2, both coming to the correct solution.

One of the two models is more capable than the other, and your task is to identify which is the more capable model (1 or 2). You MUST follow this reply format:

EXPLANATION: <Your explanation of which response was produced by the more capable model and why. 2-4 sentences.>
ANSWER: <1 or 2, indicating which response was produced by the more capable model.>
\end{promptbox}

\begin{promptbox}[colframe=pairwisecol, colbacktitle=pairwisecol!10]{Pairwise judge: user prompt}
Input question:
---

<question>

---

Multiple choice options as shown to Model 1:
A) ...

Transcript 1:
---

<transcript 1>

---

Multiple choice options as shown to Model 2:
A) ...

Transcript 2:
---

<transcript 2>

---

Which of the two transcripts is from the more capable model? Answer in the instructed format.
\end{promptbox}

Unparsable responses are treated as missing. The main condition uses complete transcripts, including visible reasoning. Comparisons that exceed the judge's context window are omitted rather than truncated.

\paragraph{Aggregating comparisons.}
To aggregate the pairwise win/loss comparisons into a single prediction of the capability score of the target model, we use a weighted average with inverse precision scaling, adapted from \citet{randomeffect2009, bradlyterry}.

Let $w_{b}^{(m,o)}$ denote the number of times model $m$ is preferred over opponent $o$ on benchmark $b$, out of $n_{b}^{(m,o)}$ judgements.
Then the empirical log odds of model $m$ winning against model $o$ is:
\[
 \operatorname{logit}\left(P(m\succ o)\right) \approx
\ell_{b}^{(m,o)}
=
\log
\frac{w_{b}^{(m,o)}+1}
{n_{b}^{(m,o)}-w_{b}^{(m,o)}+1}.
\]

The +1 terms smooth the estimate and prevent infinite log odds.

For each benchmark $b$, we calibrate the relationship between judge preference and capability difference using reference--reference comparisons:
\[
\ell_{b}^{(r,r')}
=
k_b\cdot(\theta_r^*-\theta_{r'}^*).
\]
This represents the relationship between the ground-truth difference of capability and how often a judge actually picks the more capable model.
In this relation, $k_b$ is the \textit{sensitivity} of the judge to differences in capability. Higher sensitivity means that the judge will pick the better model more frequently.

We fit $k_b$ as the slope of an ordinary least squares fit of $\ell_b^{(r,r')}$ on $\theta_r^*-\theta_{r'}^*$. The fit includes an intercept which we discard: since $\ell_b^{(r,r')}=-\ell_b^{(r',r)}$, the true intercept is zero.

Benchmarks with insufficient reference comparisons or a non-positive fitted sensitivity $k_b$ are excluded.
Given a single benchmark $b$ and reference $r$, the estimate of the target capability is:
\[
\hat{\theta}_{m|b,r} = \theta_r^* + \frac{\ell_b^{(m,r)}}{k_b},
\]
and the variance of this estimate across reference models is:
\[
\mathrm{Var}_r\left(\hat{\theta}_{m|b,r}\right) = \frac{\tau_b^2}{k_b^2}
\quad\Longrightarrow\quad
\text{precision } \omega_b = \frac{k_b^2}{\tau_b^2}.
\]

We estimate the residual variance $\tau_b^2$ from the reference--reference calibration and combine the target--reference comparisons using a weighted average with a Gaussian prior centred on the mean and variance of all reference models $\mathcal{N}(\bar\theta_\text{ref}, s^2_\text{ref})$. The average is weighted by the precision parameter $\omega_b$ we just estimated.

\[
\hat{\theta}_m
=
\frac{
\sum_{(b,r)}
\omega_b \cdot
\hat{\theta}_{m|b,r}
+
\bar{\theta}_{\mathrm{ref}}/s_{\mathrm{ref}}^2
}{
\sum_{(b,r)}
\omega_b
+
1/s_{\mathrm{ref}}^2
},
\]
where $s_{\mathrm{ref}}^2$ is the sample variance of reference capabilities. A target with no usable comparisons receives the prior mean, which is equal to the baseline prediction.

\paragraph{Leave-one-model-out calibration.}
For each prediction task, the three highest-capability reference models are considered as candidate judges.
To evaluate each candidate judge, we remove one reference model at a time, recalibrate the protocol on the remaining references, and predict the capability of the held-out model. We select the judge with the lowest mean squared leave-one-model-out error across the reference models.
Ablations on this are available in Appendix~\ref{app:judge_choice}.

\paragraph{Sensitivity and ablations.}
We conduct extensive sensitivity analyses to examine how protocol performance depends on judge aggregation, evaluation budget, and reference calibration.
First, combining judgements across a panel of models gives little benefit over a single well-selected judge (Appendix~\ref{app:protocols:ensemble}).
Second, scaling the number of judged items provides diminishing returns, as performance is often bounded by systematic judge biases rather than finite-sample noise (Appendix~\ref{app:item_abl}).
Third, this protocol is notably more expensive than others, and we analyse the trade-off between cost and prediction accuracy in Appendix~\ref{app:cost}.
Finally, calibration quality is highly dependent on the composition of the reference set: anchoring predictions requires high-capability reference models, particularly when evaluating frontier targets (Appendix~\ref{app:reference_item_abl}).

\subsection{Single-Transcript Absolute Scoring}
\label{app:protocols:direct}

\paragraph{Ratings.}
The judge receives one correct transcript at a time and assigns a capability score from 1 to 10. Candidate judge models, temperature, transcript representation, and reasoning-access condition match those used for pairwise judging. The prompts are reproduced verbatim; the options block appears only for multiple-choice tasks.

\begin{promptbox}[colframe=directcol, colbacktitle=directcol!10]{Absolute judge: system prompt}
You are an expert LLM evaluator, tasked with judging the capability of an AI model from its response to a task.
You will be given the question and one response, labelled Transcript 1, which comes to the correct solution.

For multiple choice questions, interpret any answer letter using the options as shown to this model, listed immediately before the transcript.

Estimate the model's capabilities.

Rate capability on a scale from 1 to 10, where 1 indicates very low capability (worse than the average human), 3 is the average person, 6 is on par with the top 1% of all humans, and 7-10 indicates very high capabilities beyond human experts.

You MUST follow this reply format:

EXPLANATION: <Explain the capability demonstrated by the model and why, in 2-4 sentences.>
ANSWER: <One number from 1 to 10, indicating the model's capability.>
\end{promptbox}

\begin{promptbox}[colframe=directcol, colbacktitle=directcol!10]{Absolute judge: user prompt}
Input question:
---

<question>

---

Multiple choice options as shown to Model 1:
A) ...

Transcript 1:
---

<transcript>

---

How capable is the model that produced this transcript, on a scale from 1 to 10? Answer in the instructed format.
\end{promptbox}

Unparsable responses and ratings outside the 1--10 range are treated as missing.

\paragraph{Calibration.}
Let $\bar{s}_m$ denote the mean transcript rating for model $m$. We fit
\[
\theta_r^*
=
\gamma_0
+
\gamma_1 \bar{s}_r
\]
using the reference models and predict target capability as
\[
\hat{\theta}_m
=
\gamma_0
+
\gamma_1 \bar{s}_m.
\]
The judge is selected using leave-one-model-out prediction error on the reference models, following the same procedure as pairwise judging.

\subsection{Deterministic Transcript Features}
\label{app:protocols:features}

Each transcript is represented using deterministic features derived from model-written text and tool-use metadata. These include transcript length, visible-reasoning length, the presence of visible reasoning, tool-call count, and lexical rates capturing pronouns, hedging, backtracking, uncertainty, confidence, questions, verification language, causal and inferential connectives, and contrast markers. Lexical rates are computed per 1,000 words after removing code, mathematical spans, and final-answer lines.

\begin{table}[h]
\centering
\small
\caption{\textbf{Deterministic transcript features.}}
\label{tab:det-features}
\begin{tabular}{lp{0.62\linewidth}}
\toprule
\textbf{Feature} & \textbf{Definition} \\
\midrule
Length & Log model-written transcript length \\
Reasoning length & Log visible-reasoning length \\
Visible reasoning & Indicator for non-empty visible reasoning \\
Personal pronouns & Rate of first-person pronouns \\
Hedging & Rate of hedging expressions \\
Backtracking & Rate of self-correction expressions \\
Uncertainty & Rate of uncertainty expressions \\
Confidence & Rate of confidence expressions \\
Questions & Question-mark rate \\
Verification & Rate of checking and verification expressions \\
Connectives & Rate of causal and inferential connectives \\
Contrast & Rate of contrastive expressions \\
Tool calls & Log number of tool calls \\
\bottomrule
\end{tabular}
\end{table}

\paragraph{Regression.}
For each model, we average transcript features across its correctly solved sampled items to obtain a model-level representation. Features are standardised using the reference models, and ridge regression is fitted to predict $\theta^*$. The regularisation coefficient is selected by leave-one-model-out cross-validation over 13 logarithmically spaced values from $10^{-3}$ to $10^3$.

\subsection{Transcript Embeddings}
\label{app:protocols:embeddings}

We embed each rendered transcript, including visible reasoning when available, using Qwen3-Embedding-8B at a fixed model revision. Transcripts that exceed the encoder context window are divided into overlapping chunks of at most 8,192 tokens. Chunk embeddings are combined using a token-weighted mean and normalised to produce one representation per transcript.

For each model, we average its transcript embeddings across correctly solved sampled items to obtain a model-level representation. We then fit ridge regression from this representation to $\theta^*$ using the reference models. The regularisation coefficient is selected by leave-one-model-out cross-validation over 13 logarithmically spaced values from $10^{-2}$ to $10^4$. The main results use ridge regression on full-transcript embeddings.

\subsection{Worked Example: Pairwise and Absolute Judging}
\label{app:protocols:example}

We show one GPQA-Diamond item from the \science{} 2026-I prediction task, answered correctly by the reference model gpt-oss-20b ($\theta^* = 0.01$) and the target model glm-5.2 ($\theta^* = 1.71$). Both protocols use the same judge, kimi-k2-thinking, with visible reasoning included, and the prompts of Appendix~\ref{app:protocols:pairwise} and~\ref{app:protocols:direct}. Transcripts and judge outputs are reproduced verbatim except that Unicode symbols are typeset as mathematics; each judge box shows the judge's reasoning (grey) above the explanation and verdict it returned.

\begin{exbox}[colupper=black]{GPQA-Diamond item}
Consider a system with Hamiltonian operator $H = \varepsilon \vec{\sigma}.\vec{n}$. Here, $\vec{n}$ is an arbitrary unit vector, $\varepsilon$ is a constant of dimension energy, and components of $\vec{\sigma}$ are the Pauli spin matrices. What are the eigenvalues of the Hamiltonian operator?

\smallskip
Options shown to both models: A) $+\varepsilon, -\varepsilon$ \quad B) $+\hbar/2, -\hbar/2$ \quad C) $+1, -1$ \quad D) $+\varepsilon\hbar/2, -\varepsilon\hbar/2$
\end{exbox}

\begin{exbox}{gpt-oss-20b \textmd{(reference, $\theta^*=0.01$, 722 characters)}}
\boxlabel{Reasoning}
We need eigenvalues of $H = \varepsilon (\sigma \cdot n)$. For any unit vector $n$, $\sigma \cdot n$ has eigenvalues $\pm 1$. Because $\sigma\cdot n = (n\cdot\sigma)$ has eigenvalues $\pm 1$. Then $H$ has eigenvalues $\pm\varepsilon$. So answer is A.

But double-check: The $\sigma\cdot n$ is a 2x2 hermitian matrix with eigenvalues $\pm 1$. Classic. So $H$ eigenvalues are $\pm\varepsilon$.

Given choices:

A) $+\varepsilon, -\varepsilon$. B $+\hbar/2, -\hbar/2$. C $+1, -1$. D $+\varepsilon\hbar/2, -\varepsilon\hbar/2$. So A.

Thus answer: A.

Now produce final.
\tcblower
\boxlabel{Answer}
The operator $\vec{\sigma}\cdot\vec{n}$ is a $2\times2$ Hermitian matrix whose eigenvalues are $\pm 1$ (since $(\vec{\sigma}\cdot\vec{n})^2=\mathbb{I}$). Multiplying by the scalar $\varepsilon$ simply scales these eigenvalues, giving eigenvalues $+\varepsilon$ and $-\varepsilon$.

ANSWER: A
\end{exbox}

\begin{exbox}{glm-5.2 \textmd{(target, $\theta^*=1.71$, 2{,}141 characters)}}
\boxlabel{Reasoning}
The Hamiltonian is $H = \varepsilon \vec{\sigma} \cdot \vec{n}$, where $\vec{n}$ is a unit vector.\\
The Pauli spin matrices satisfy the property that for any unit vector $\vec{n}$, the eigenvalues of $\vec{\sigma} \cdot \vec{n}$ are +1 and -1.\\
Proof: $(\vec{\sigma} \cdot \vec{n})^2 = \sigma_i n_i \sigma_j n_j = \frac{1}{2} (\sigma_i \sigma_j + \sigma_j \sigma_i) n_i n_j = \delta_{ij} n_i n_j = n_i n_i = |\vec{n}|^2 = 1$.\\
Since the trace of $\vec{\sigma} \cdot \vec{n}$ is 0, the sum of eigenvalues is 0.\\
Since the square has eigenvalues 1, the eigenvalues must be 1 and -1.\\
Therefore, the eigenvalues of $H = \varepsilon \vec{\sigma} \cdot \vec{n}$ are $\varepsilon \times 1$ and $\varepsilon \times (-1)$, which are $+\varepsilon$ and $-\varepsilon$.

Let's check the options:\\
A) $+\varepsilon, -\varepsilon$\\
B) $+\hbar/2, -\hbar/2$\\
C) $+1, -1$\\
D) $+\varepsilon \hbar/2, - \varepsilon \hbar/2$

Option A matches perfectly.
\tcblower
\boxlabel{Answer}
The Hamiltonian is given by $H = \varepsilon \vec{\sigma} \cdot \vec{n}$, where $\vec{n}$ is a unit vector and $\vec{\sigma}$ represents the Pauli spin matrices.

To find the eigenvalues of $H$, we first determine the eigenvalues of the operator $\vec{\sigma} \cdot \vec{n}$.\\
A well-known property of the Pauli matrices is that for any arbitrary unit vector $\vec{n}$, the operator $\vec{\sigma} \cdot \vec{n}$ has eigenvalues $+1$ and $-1$. This can be seen because:\\
1. The trace of $\vec{\sigma} \cdot \vec{n}$ is zero, meaning the sum of its eigenvalues is 0.\\
2. $(\vec{\sigma} \cdot \vec{n})^2 = \sum_{i,j} \sigma_i \sigma_j n_i n_j = \sum_{i,j} (\delta_{ij} + i \epsilon_{ijk} \sigma_k) n_i n_j = \sum_i n_i^2 + i \sum_{i,j,k} \epsilon_{ijk} n_i n_j \sigma_k = |\vec{n}|^2 + 0 = 1$.\\
Since the sum of the eigenvalues is 0 and the sum of their squares is 1, the eigenvalues must be $+1$ and $-1$.

Therefore, the eigenvalues of the Hamiltonian $H = \varepsilon \vec{\sigma} \cdot \vec{n}$ are obtained by multiplying the eigenvalues of $\vec{\sigma} \cdot \vec{n}$ by $\varepsilon$:\\
$\lambda_1 = \varepsilon (+1) = +\varepsilon$\\
$\lambda_2 = \varepsilon (-1) = -\varepsilon$

ANSWER: A
\end{exbox}

\paragraph{Pairwise judge.}
The pair is judged in both orders (Appendix~\ref{app:protocols:pairwise}); the judge prefers glm-5.2 in both.

\begin{exbox}[colframe=pairwisecol, colbacktitle=pairwisecol!10, fontupper=\footnotesize, after title={\hfill Verdict: glm-5.2}]{Order 1: \textmd{gpt-oss-20b = Transcript 1, glm-5.2 = Transcript 2}}
\boxlabel{Judge reasoning}
The question: $H = \varepsilon \vec{\sigma} \cdot \vec{n}$, find eigenvalues. Both models correctly identify that $\vec{\sigma} \cdot \vec{n}$ has eigenvalues $\pm 1$, so $H$ has eigenvalues $\pm \varepsilon$. Both select option A. Now comparing the quality of reasoning:

Transcript 1: Brief thought process; recognizes that $\sigma \cdot n$ has eigenvalues $\pm 1$ because it's a standard result. Mentions that $(\vec{\sigma}\cdot\vec{n})^2 = \mathbb{I}$ (implied by ``eigenvalues are ±1''). Quick verification of options. Final answer is concise and correct.

Transcript 2: More detailed derivation. Shows the proof that $(\vec{\sigma} \cdot \vec{n})^2 = 1$. Uses the trace argument (sum of eigenvalues is 0). Explicitly calculates $(\vec{\sigma} \cdot \vec{n})^2$ using Pauli matrix algebra: $\sigma_i \sigma_j = \delta_{ij} + i\epsilon_{ijk}\sigma_k$. Then concludes eigenvalues are $\pm 1$. Then scales by $\varepsilon$.

Transcript 1 is brief but correct; it relies on ``classic'' knowledge. Transcript 2 provides a rigorous derivation from first principles (Pauli matrix commutation relations) and explicitly mentions the trace condition. In an evaluation of capability, demonstrating the ability to derive from first principles is generally considered a higher level of capability than just recalling a fact, assuming both are correct. Transcript 2 also explicitly mentions the trace condition and the squaring property, which are the key ingredients.

Therefore, Transcript 2 is from the more capable model.
\tcblower
\boxlabel{Explanation}
Transcript 2 provides a rigorous, step-by-step derivation of the eigenvalues from the algebraic properties of the Pauli matrices, explicitly proving that $(\vec{\sigma}\cdot\vec{n})^2 = \mathbb{I}$ and using the trace condition to confirm the eigenvalues are $\pm 1$ before scaling by $\varepsilon$. In contrast, Transcript 1 relies on recalling the result as a ``classic'' property without derivation. While both arrive at the correct answer, the ability to derive the result from first principles demonstrates deeper physical and mathematical understanding.

\smallskip\textbf{ANSWER: 2}
\end{exbox}

\begin{exbox}[colframe=pairwisecol, colbacktitle=pairwisecol!10, fontupper=\footnotesize, after title={\hfill Verdict: glm-5.2}]{Order 2: \textmd{glm-5.2 = Transcript 1, gpt-oss-20b = Transcript 2}}
\boxlabel{Judge reasoning}
Comparing two transcripts from two models answering a physics question about Hamiltonian eigenvalues. Both models correctly identify that $\vec{\sigma} \cdot \vec{n}$ has eigenvalues $\pm 1$, making the Hamiltonian's eigenvalues $\pm \varepsilon$, and both select option A.

Transcript 1: Demonstrates thorough command of the formalism. Explicitly proves $(\vec{\sigma}\cdot\vec{n})^2 = \mathbb{I}$ using index notation. Invokes the traceless condition to rigorously fix the spectrum as $\{+1,-1\}$ rather than merely allowing eigenvalues $\pm 1$ individually. Uses the calculation: $(\vec{\sigma} \cdot \vec{n})^2 = \sigma_i n_i \sigma_j n_j = \frac{1}{2} (\sigma_i \sigma_j + \sigma_j \sigma_i) n_i n_j = \delta_{ij} n_i n_j = n_i n_i = |\vec{n}|^2 = 1$. Since the trace of $\vec{\sigma} \cdot \vec{n}$ is 0, the sum of eigenvalues is 0. Since the square has eigenvalues 1, the eigenvalues must be 1 and -1.

Transcript 2: Reaches the correct answer with less detail. Omits the trace argument needed to fully exclude degenerate spectra. More concise but less rigorous.

Transcript 1's explanation is more complete and educational. It shows the mathematical steps clearly using index notation and explicitly mentions both the trace condition and the squaring property. The logic is sound and the derivation is complete.

Therefore, Transcript 1 is from the more capable model.
\tcblower
\boxlabel{Explanation}
Transcript 1 demonstrates a more thorough and rigorous command of the formalism. It explicitly proves $(\vec{\sigma}\cdot\vec{n})^2 = \mathbb{I}$ using index notation and, crucially, invokes the traceless condition to rigorously fix the spectrum as $\{+1,-1\}$ rather than merely allowing eigenvalues $\pm 1$ individually. Transcript 2 reaches the correct answer with less detail and omits the trace argument needed to fully exclude degenerate spectra.

\smallskip\textbf{ANSWER: 1}
\end{exbox}

\paragraph{Absolute judge.}
Rated one at a time (Appendix~\ref{app:protocols:direct}), the two transcripts receive the same score, so this item carries no signal for absolute scoring while it separates the two models under pairwise judging.

\begin{exbox}[colframe=directcol, colbacktitle=directcol!10, fontupper=\footnotesize, after title={\hfill Verdict: 6}]{gpt-oss-20b}
\boxlabel{Judge reasoning}
Evaluating the capability of an AI model based on its response to a physics question about quantum mechanics. The question asks for the eigenvalues of the Hamiltonian $H = \varepsilon \vec{\sigma} \cdot \vec{n}$, where $\vec{n}$ is a unit vector and $\vec{\sigma}$ are the Pauli matrices.

The model correctly identifies that:
\begin{enumerate}
\item $\vec{\sigma} \cdot \vec{n}$ is a spin operator along direction $\vec{n}$
\item The eigenvalues of $\vec{\sigma} \cdot \vec{n}$ are $\pm 1$ (since $(\vec{\sigma} \cdot \vec{n})^2 = I$)
\item Therefore, the eigenvalues of $H = \varepsilon (\vec{\sigma} \cdot \vec{n})$ are $\pm \varepsilon$
\item This corresponds to option A
\end{enumerate}

The reasoning is correct and concise. The model shows understanding of:
\begin{itemize}
\item Pauli matrices properties
\item Spin operators in quantum mechanics
\item Eigenvalue scaling under scalar multiplication
\item The relationship between the dot product of Pauli matrices with a unit vector and the resulting operator's eigenvalues
\end{itemize}

This is a standard quantum mechanics problem, typically taught in undergraduate or early graduate quantum mechanics courses. Getting this right demonstrates solid understanding of quantum mechanics concepts.

The model's explanation is clear, concise, and correct. It shows understanding of why the eigenvalues are what they are, not just memorization. The operator $\vec{\sigma}\cdot\vec{n}$ is correctly identified as a $2\times2$ Hermitian matrix whose eigenvalues are ±1 (since $(\vec{\sigma}\cdot\vec{n})^2=\mathbb{I}$), and the scaling by $\varepsilon$ is correct.

Rating: 6 (on par with top 1\% of humans --- those who have studied quantum mechanics).
\tcblower
\boxlabel{Explanation}
The model correctly identifies that the operator $\vec{\sigma}\cdot\vec{n}$ (representing spin along direction $\vec{n}$) has eigenvalues $\pm 1$ since $(\vec{\sigma}\cdot\vec{n})^2 = \mathbb{I}$ for any unit vector $\vec{n}$, and correctly applies scalar multiplication to obtain eigenvalues $\pm\varepsilon$. This demonstrates solid understanding of quantum spin physics and linear algebra concepts typically taught in undergraduate quantum mechanics courses.

\smallskip\textbf{ANSWER: 6}
\end{exbox}

\begin{exbox}[colframe=directcol, colbacktitle=directcol!10, fontupper=\footnotesize, after title={\hfill Verdict: 6}]{glm-5.2}
\boxlabel{Judge reasoning}
Evaluating the capability of the AI model that produced this transcript in response to the physics question about Hamiltonian eigenvalues.

First, analyze the response: The model correctly identifies that we need to find the eigenvalues of $\vec{\sigma} \cdot \vec{n}$ first. It correctly states that the eigenvalues of $\vec{\sigma} \cdot \vec{n}$ are +1 and -1. It provides a rigorous proof:
\begin{itemize}
\item Uses the property that $(\vec{\sigma} \cdot \vec{n})^2 = 1$ (since $\vec{n}$ is a unit vector and Pauli matrices satisfy $\sigma_i \sigma_j = \delta_{ij} + i\epsilon_{ijk}\sigma_k$).
\item Correctly notes that the trace is 0, so eigenvalues sum to 0.
\item Correctly concludes that since $(\vec{\sigma} \cdot \vec{n})^2 = I$, the eigenvalues squared are 1, so they are $\pm 1$.
\item Then correctly scales by $\varepsilon$ to get eigenvalues $\pm \varepsilon$.
\item Correctly identifies option A as the answer.
\end{itemize}

The reasoning is sound, mathematically correct, and physically accurate. This is a standard result in quantum mechanics (spin-1/2 systems), related to the fact that $\vec{\sigma} \cdot \vec{n}$ represents spin projection along direction $\vec{n}$, which has eigenvalues $\pm 1$ (for Pauli matrices specifically).

The model not only solves it but provides clear reasoning, checks the trace condition, verifies the square of the operator, and correctly maps to the multiple choice options. The solution is complete and rigorous.

The model provides a perfect solution with clear mathematical derivation. This suggests capability around 6. It's not ``beyond human experts'' because this is a basic problem that any physics expert would know, but it's certainly better than the average human. An average person wouldn't know this, and an undergraduate physics student should. A graduate student in physics definitely knows this.

Rating: 6 (on par with top 1\% of humans --- those who have studied quantum mechanics).
\tcblower
\boxlabel{Explanation}
The model correctly solves the quantum mechanics problem by recognizing that the operator $\vec{\sigma} \cdot \vec{n}$ represents spin projection along the unit vector $\vec{n}$ and has eigenvalues $\pm 1$. It rigorously derives this using the Pauli matrix identity $(\vec{\sigma} \cdot \vec{n})^2 = |\vec{n}|^2 = 1$ and the traceless property, then correctly scales by $\varepsilon$ to obtain eigenvalues $\pm \varepsilon$, matching option A. This demonstrates accurate physical reasoning and mathematical competence at the level of a trained physicist.

\smallskip\textbf{ANSWER: 6}
\end{exbox}

None of the four verdicts notices a slip in glm-5.2's answer: it states that ``the sum of their squares is 1'', whereas the squared eigenvalues sum to $\operatorname{tr}(\vec{\sigma}\cdot\vec{n})^2 = \operatorname{tr}\mathbb{I} = 2$. The pairwise preference rests on the presence of an explicit derivation rather than on its correctness, and the absolute judge gives both transcripts the same score.

\clearpage

\section{Additional Results}

\subsection{Protocol performance}
\label{app:results_signal}
\begin{table}[!h]
    \label{tab:results_signal}
    \centering
    \small
    \caption{
        \textbf{Prediction error per protocol.} \textit{(Accompaniment to Figure~\ref{fig:signal-capability}.)}
        \textit{MSE}: Mean squared error over the 16 target models of late 2025 and early 2026 (averaged over four domains). $\Delta$SE: mean squared error minus the baseline's, paired $t$-test ($n=16$ target models). \textit{LLR}: mean log likelihood ratio of the protocol's prediction against the baseline protocol (standard error in parentheses). $p$: one-sample $t$-test on LLR against 0. We report statistics based on LLR in Section~\ref{sec:results-signal}, to penalise confidently incorrect predictions.
    }
    \label{tab:signal-capability}
    \begin{tabular}{lccccc}
        \toprule
        \textbf{Protocol} & \textbf{MSE [95\% CI]} & \textbf{$\Delta$SE} & \textbf{$p$} & \textbf{LLR} & \textbf{$p$} \\
        \midrule
        Baseline protocol & 1.37 [0.88, 1.87] & -- & -- & -- & -- \\
        Deterministic features & 0.86 [0.52, 1.21] & $-0.51$ & .064 & $-0.26$ (0.30) & .404 \\
        Embedding & 1.01 [0.61, 1.41] & $-0.36$ & $<$.001 & 0.04 (0.06) & .464 \\
        Absolute scoring & 1.01 [0.67, 1.35] & $-0.36$ & .002 & $-0.21$ (0.16) & .200 \\
        Pairwise judge & 0.26 [0.18, 0.34] & $-1.12$ & $<$.001 & 0.68 (0.21) & .005 \\
        \bottomrule
    \end{tabular}
\end{table}

\subsection{Prediction performance across historical target periods}
\label{app:back-testing}
\begin{table}[!h]
    \centering
    \small
    \caption{
        \textbf{Prediction error per protocol across target periods.}
        \textit{Accompaniment to Figure~\ref{fig:time_generalisation}.}
        Mean squared error over the 8 target models of each period, with 95\% percentile-bootstrap intervals (2,000 resamples). Lowest error per row underlined. Cell colours show mean squared error on a logarithmic scale from 0.1 (blue) to 10+ (red).
    }
    \label{tab:time-generalisation}
    \resizebox{\textwidth}{!}{%
    \begin{tabular}{llccccc}
        \toprule
        \textbf{Domain} & \textbf{Target period} & \textbf{Baseline protocol} & \textbf{Deterministic features} & \textbf{Embedding} & \textbf{Absolute scoring} & \textbf{Pairwise judge} \\
        \midrule
        \science & Early 2024 & \cellcolor{red!13}1.65 {\scriptsize [1.05, 2.27]} & \cellcolor{red!13}1.66 {\scriptsize [1.07, 2.29]} & \cellcolor{red!13}1.65 {\scriptsize [1.05, 2.25]} & \cellcolor{red!12}1.61 {\scriptsize [1.00, 2.20]} & \cellcolor{blue!58}\underline{0.11} {\scriptsize [0.01, 0.24]} \\
         & Late 2024 & \cellcolor{blue!16}0.54 {\scriptsize [0.36, 0.74]} & \cellcolor{blue!14}0.58 {\scriptsize [0.37, 0.84]} & \cellcolor{blue!16}0.54 {\scriptsize [0.37, 0.74]} & \cellcolor{blue!20}\underline{0.46} {\scriptsize [0.32, 0.63]} & \cellcolor{blue!16}0.54 {\scriptsize [0.36, 0.75]} \\
         & Early 2025 & \cellcolor{red!4}1.16 {\scriptsize [0.61, 1.72]} & \cellcolor{red!60}38.06 {\scriptsize [4.01, 82.83]} & \cellcolor{blue!7}0.77 {\scriptsize [0.38, 1.20]} & \cellcolor{red!37}4.14 {\scriptsize [1.52, 6.84]} & \cellcolor{blue!11}\underline{0.65} {\scriptsize [0.24, 1.08]} \\
         & Late 2025 & \cellcolor{blue!4}0.85 {\scriptsize [0.46, 1.23]} & \cellcolor{blue!33}0.28 {\scriptsize [0.14, 0.46]} & \cellcolor{blue!21}0.45 {\scriptsize [0.26, 0.66]} & \cellcolor{blue!23}0.42 {\scriptsize [0.24, 0.63]} & \cellcolor{blue!46}\underline{0.17} {\scriptsize [0.02, 0.37]} \\
         & Early 2026 & \cellcolor{blue!21}0.45 {\scriptsize [0.27, 0.64]} & \cellcolor{blue!15}0.57 {\scriptsize [0.18, 1.11]} & \cellcolor{blue!27}0.35 {\scriptsize [0.18, 0.54]} & \cellcolor{blue!42}\underline{0.20} {\scriptsize [0.06, 0.37]} & \cellcolor{blue!36}0.25 {\scriptsize [0.13, 0.38]} \\
        \midrule
        \maths & Early 2024 & \cellcolor{red!12}1.61 {\scriptsize [0.98, 2.25]} & \cellcolor{red!18}1.97 {\scriptsize [0.81, 3.13]} & \cellcolor{red!9}\underline{1.42} {\scriptsize [0.78, 2.14]} & \cellcolor{red!23}2.41 {\scriptsize [1.19, 3.76]} & \cellcolor{red!12}1.61 {\scriptsize [0.97, 2.23]} \\
         & Late 2024 & \cellcolor{blue!17}0.53 {\scriptsize [0.22, 0.92]} & \cellcolor{blue!23}0.42 {\scriptsize [0.09, 0.84]} & \cellcolor{blue!27}0.35 {\scriptsize [0.14, 0.62]} & \cellcolor{blue!32}0.29 {\scriptsize [0.07, 0.54]} & \cellcolor{blue!58}\underline{0.11} {\scriptsize [0.02, 0.22]} \\
         & Early 2025 & \cellcolor{red!32}3.41 {\scriptsize [1.95, 4.95]} & \cellcolor{red!60}578.38 {\scriptsize [71.89, 1377.78]} & \cellcolor{red!25}\underline{2.59} {\scriptsize [1.30, 3.89]} & \cellcolor{red!29}3.03 {\scriptsize [1.56, 4.53]} & \cellcolor{red!36}3.97 {\scriptsize [1.76, 6.10]} \\
         & Late 2025 & \cellcolor{red!21}2.23 {\scriptsize [1.21, 3.24]} & \cellcolor{blue!3}0.89 {\scriptsize [0.40, 1.41]} & \cellcolor{red!1}1.04 {\scriptsize [0.50, 1.57]} & \cellcolor{red!11}1.51 {\scriptsize [0.98, 2.02]} & \cellcolor{blue!34}\underline{0.27} {\scriptsize [0.07, 0.51]} \\
         & Early 2026 & \cellcolor{blue!23}0.41 {\scriptsize [0.24, 0.58]} & \cellcolor{blue!12}0.63 {\scriptsize [0.26, 1.05]} & \cellcolor{blue!32}0.29 {\scriptsize [0.18, 0.40]} & \cellcolor{blue!42}\underline{0.20} {\scriptsize [0.08, 0.34]} & \cellcolor{blue!15}0.57 {\scriptsize [0.30, 0.85]} \\
        \midrule
        \code & Late 2025 & \cellcolor{red!25}2.65 {\scriptsize [1.18, 3.91]} & \cellcolor{blue!31}0.30 {\scriptsize [0.09, 0.57]} & \cellcolor{red!25}2.65 {\scriptsize [1.28, 3.94]} & \cellcolor{red!18}2.03 {\scriptsize [1.05, 2.97]} & \cellcolor{blue!46}\underline{0.17} {\scriptsize [0.01, 0.43]} \\
         & Early 2026 & \cellcolor{red!13}1.64 {\scriptsize [1.03, 2.34]} & \cellcolor{red!2}1.09 {\scriptsize [0.49, 1.73]} & \cellcolor{blue!8}0.74 {\scriptsize [0.40, 1.11]} & \cellcolor{blue!6}0.78 {\scriptsize [0.39, 1.25]} & \cellcolor{blue!58}\underline{0.11} {\scriptsize [0.05, 0.20]} \\
        \midrule
        \cyber & Late 2025 & \cellcolor{red!2}1.08 {\scriptsize [0.48, 1.75]} & \cellcolor{blue!5}0.82 {\scriptsize [0.27, 1.48]} & \cellcolor{red!2}1.08 {\scriptsize [0.50, 1.76]} & \cellcolor{red!0}1.01 {\scriptsize [0.45, 1.82]} & \cellcolor{blue!30}\underline{0.32} {\scriptsize [0.06, 0.69]} \\
         & Early 2026 & \cellcolor{red!14}1.70 {\scriptsize [1.16, 2.34]} & \cellcolor{red!22}2.31 {\scriptsize [1.41, 3.33]} & \cellcolor{red!10}1.49 {\scriptsize [0.98, 2.09]} & \cellcolor{red!17}1.95 {\scriptsize [1.28, 2.75]} & \cellcolor{blue!42}\underline{0.20} {\scriptsize [0.08, 0.33]} \\
        \bottomrule
    \end{tabular}}
\end{table}

\subsection{Specific choice of judge model does not affect prediction performance}
\label{app:judge_choice}

\begin{table*}[!h]
  \centering
  \caption{
    \textbf{Candidate and picked judges.}
    For each target semester and domain, the three references (models from the preceding semester) with the highest global capability $\theta$, which serve as candidate judges for pairwise judging and absolute scoring, ordered by reference leave-one-out error. The candidate with the lowest error on the reference set (``\textit{Best}'') is the judge used for the task.
  }
  \label{tab:loo-judges}
  \small
  \resizebox{\textwidth}{!}{
  \begin{tabular}{lllccc}
    \toprule
    \textbf{Semester} & \textbf{Domain} & \textbf{Protocol} & \textbf{Best} & \textbf{2nd Best} & \textbf{3rd Best} \\
    \midrule
    \multirow{4}{*}{2024-I}
     & \multirow{2}{*}{\science} & Pairwise & \textit{gpt-4-0613} & mixtral-8x7b-instruct-v0.1 & nous-hermes-2-yi-34b \\
     &  & Absolute & nous-hermes-2-yi-34b & \textit{gpt-4-0613} & mixtral-8x7b-instruct-v0.1 \\
    \cmidrule{2-6}
     & \multirow{2}{*}{\maths} & Pairwise & \textit{gpt-3.5-turbo-0125} & mixtral-8x7b-instruct-v0.1 & \textit{gpt-4-0613} \\
     &  & Absolute & \textit{gpt-3.5-turbo-0125} & \textit{gpt-4-0613} & mixtral-8x7b-instruct-v0.1 \\
    \midrule
    \multirow{4}{*}{2024-II}
     & \multirow{2}{*}{\science} & Pairwise & \textit{gpt-4o-2024-05-13} & \textit{gpt-4-turbo-2024-04-09} & wizardlm-2-8x22b \\
     &  & Absolute & \textit{gpt-4o-2024-05-13} & \textit{gpt-4-turbo-2024-04-09} & wizardlm-2-8x22b \\
    \cmidrule{2-6}
     & \multirow{2}{*}{\maths} & Pairwise & \textit{gpt-4-turbo-2024-04-09} & \textit{gpt-4o-2024-05-13} & qwen2-72b-instruct \\
     &  & Absolute & \textit{gpt-4-turbo-2024-04-09} & qwen2-72b-instruct & \textit{gpt-4o-2024-05-13} \\
    \midrule
    \multirow{4}{*}{2025-I}
     & \multirow{2}{*}{\science} & Pairwise & qwen2.5-72b-instruct & deepseek-v3 & phi-4 \\
     &  & Absolute & phi-4 & qwen2.5-72b-instruct & deepseek-v3 \\
    \cmidrule{2-6}
     & \multirow{2}{*}{\maths} & Pairwise & llama-3.3-70b-instruct & deepseek-v3 & phi-4 \\
     &  & Absolute & phi-4 & deepseek-v3 & llama-3.3-70b-instruct \\
    \midrule
    \multirow{8}{*}{2025-II}
     & \multirow{2}{*}{\science} & Pairwise & minimax-m1 & deepseek-r1-0528 & qwen3-235b-a22b \\
     &  & Absolute & deepseek-r1-0528 & minimax-m1 & qwen3-235b-a22b \\
    \cmidrule{2-6}
     & \multirow{2}{*}{\maths} & Pairwise & deepseek-r1-0528 & qwen3-235b-a22b & minimax-m1 \\
     &  & Absolute & qwen3-235b-a22b & minimax-m1 & deepseek-r1-0528 \\
    \cmidrule{2-6}
     & \multirow{2}{*}{\code} & Pairwise & minimax-m1 & deepseek-r1-0528 & qwen3-235b-a22b \\
     &  & Absolute & qwen3-235b-a22b & deepseek-r1-0528 & minimax-m1 \\
    \cmidrule{2-6}
     & \multirow{2}{*}{\cyber} & Pairwise & deepseek-r1-0528 & qwen3-235b-a22b & minimax-m1 \\
     &  & Absolute & minimax-m1 & qwen3-235b-a22b & deepseek-r1-0528 \\
    \midrule
    \multirow{8}{*}{2026-I}
     & \multirow{2}{*}{\science} & Pairwise & minimax-m2 & glm-4.7 & kimi-k2-thinking \\
     &  & Absolute & glm-4.7 & minimax-m2 & kimi-k2-thinking \\
    \cmidrule{2-6}
     & \multirow{2}{*}{\maths} & Pairwise & kimi-k2-thinking & glm-4.5 & glm-4.7 \\
     &  & Absolute & glm-4.7 & glm-4.5 & kimi-k2-thinking \\
    \cmidrule{2-6}
     & \multirow{2}{*}{\code} & Pairwise & kimi-k2-thinking & minimax-m2 & deepseek-v3.2 \\
     &  & Absolute & deepseek-v3.2 & minimax-m2 & kimi-k2-thinking \\
    \cmidrule{2-6}
     & \multirow{2}{*}{\cyber} & Pairwise & kimi-k2-thinking & deepseek-v3.2 & glm-4.7 \\
     &  & Absolute & deepseek-v3.2 & glm-4.7 & kimi-k2-thinking \\
    \bottomrule
  \end{tabular}
  }
\end{table*}

Our main analyses use one judge for each prediction task.
We typically consider the 3 strongest models of the previous reference semester, and pick as judge the one with the lowest leave-one-model-out prediction error on the other references.
Table~\ref{tab:loo-judges} shows the selected judge and other candidate judges.

We then test whether the results depend strongly on this judge-selection step. For each task, we repeat judging with each of the three strongest reference models, which are also the models used to define the saturated-item boundary. We compare squared prediction error across the 16 target models from 2025-II and 2026-I.

\begin{figure}[!htbp]
  \centering
  \includegraphics[width=0.6\textwidth]{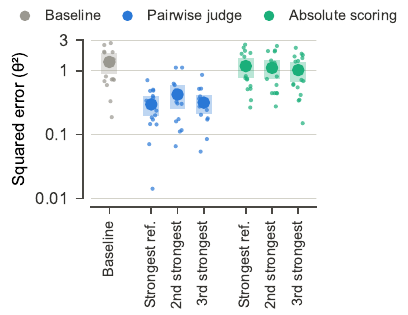}
  \caption{
  \textbf{Pairwise judging is robust to the choice of judge.}
  Squared prediction error for the 16 target models from 2025-II and 2026-I when using the strongest, second-strongest, or third-strongest reference model as judge. Small points show individual target models, averaged across domains; large circles show means and boxes show 95\% $t$-intervals.
  }
  \label{fig:judge_choice}
\end{figure}

The three judges produce similar prediction errors (Figure~\ref{fig:judge_choice}).
Changing the judge shifts a task's mean squared error by a median of 0.30, much smaller than the gap between pairwise judging and either absolute scoring (median 0.78) or the baseline (median 1.09).
The main pairwise-judging results therefore do not depend on selecting a particular judge.

\clearpage
\subsection{Pairwise judge predictions are best in \code and \cyber domains}
\label{app:per-domain}

The main analysis pools prediction error across the four domains. To test whether protocol performance varies systematically by domain, we fit a hierarchical Gamma generalised linear model with a log link function to the squared prediction errors of the 16 target models from 2025-II and 2026-I across all domains. This gives 64 target-by-domain combinations, with one squared error per protocol and combination:

\begin{equation}
\texttt{error}^2 \sim \texttt{protocol} + \texttt{domain} + (\texttt{protocol} \mid \texttt{domain}) + (1 \mid \texttt{target}:\texttt{domain})
\end{equation}

The protocol term also includes a length-only protocol (which includes only the length terms of the deterministic features; see Table~\ref{tab:det-features}), fitted jointly with the other protocols.

\begin{figure}[h]
    \centering
    \includegraphics[width=.95\textwidth]{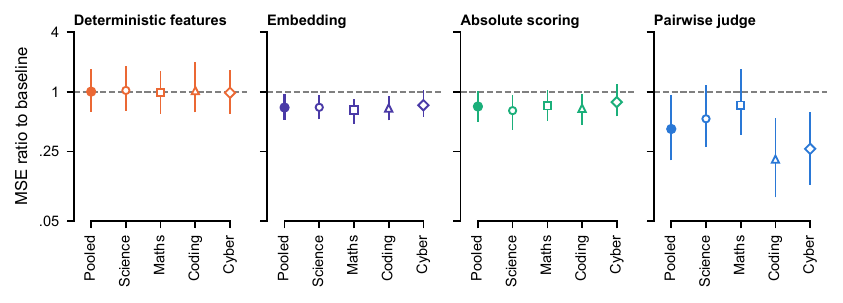}
    \caption{
        \textbf{Protocol efficacy varies across domains.}
        Posterior MSE ratio of each protocol relative to the baseline protocol, pooled across domains (\textit{filled circles}) and separately by domain (\textit{open markers}), estimated using the hierarchical Gamma GLM. Error bars show 95\% credible intervals; the dashed line marks equal error to the baseline.
    }
    \label{fig:domain_glm}
\end{figure}

We use the baseline protocol as our protocol reference, with its effect fixed at zero; domain effects are normalised to sum to zero. Protocol-by-domain deviations $(\texttt{protocol} \mid \texttt{domain})$ are partially pooled separately for each protocol and drawn from a normal distribution with protocol-specific standard deviation $\sigma_p$. This captures variation of that protocol across domains. The target-by-domain term $(1 \mid \texttt{target}:\texttt{domain})$ is a random intercept accounting for the fact that all protocols predict the same target within each domain.

We use $\mathcal{N}(0,2)$ priors for the intercept, $\mathcal{N}(0,1)$ priors for protocol and domain effects, $\operatorname{HalfNormal}(0.5)$ priors for $\sigma_p$, and a $\operatorname{HalfNormal}(1)$ prior for the standard deviation of the target-by-domain random intercepts. We fit the model using NUTS with four chains, 1,000 warm-up draws, and 2,000 retained draws per chain. The model fitting showed no divergences, with $\hat{R}\leq1.005$ and bulk effective sample size $\geq1{,}090$.

Pairwise judging has the lowest pooled MSE ratio at 0.42 (95\% credible interval 0.20--0.93), but this varies substantially by domain (Figure~\ref{fig:domain_glm}): it has clear predictive ability in \code and \cyber, but is less reliable in \science and \maths (Table~\ref{tab:main-domain-results}).

\begin{table}[h]
    \centering
    \small
    \caption{
        \textbf{Hierarchical estimates of protocol performance by domain.}
        Posterior median MSE ratio relative to the baseline protocol, with 95\% credible intervals in brackets. $\sigma_p$ shows the between-domain standard deviation.
    }
    \label{tab:main-domain-results}
    \resizebox{\textwidth}{!}{
    \begin{tabular}{lcccccc}
        \toprule
        \textbf{Protocol} & \textbf{Pooled} & \textbf{Science} & \textbf{Maths} & \textbf{Coding} & \textbf{Cybersecurity} & $\sigma_p$ \\
        \midrule
        Deterministic features & 1.00 [0.63, 1.68] & 1.03 [0.64, 1.83] & 0.98 [0.60, 1.62] & 1.03 [0.62, 2.00] & 0.98 [0.60, 1.64] & 0.17 [0.01, 0.70] \\
        Embedding & 0.69 [0.52, 0.95] & 0.70 [0.53, 0.93] & \textbf{0.65 [0.47, 0.85]} & 0.68 [0.51, 0.90] & 0.73 [0.56, 1.03] & 0.13 [0.01, 0.57] \\
        Absolute scoring & 0.71 [0.50, 1.03] & 0.65 [0.41, 0.92] & 0.72 [0.51, 1.04] & 0.68 [0.46, 0.96] & 0.79 [0.56, 1.20] & 0.18 [0.01, 0.68] \\
        Pairwise judging & \textbf{0.42 [0.20, 0.93]} & \textbf{0.53 [0.28, 1.17]} & 0.73 [0.37, 1.68] & \textbf{0.21 [0.09, 0.55]} & \textbf{0.27 [0.12, 0.63]} & 0.62 [0.12, 1.22] \\
        \bottomrule
    \end{tabular}
    }
\end{table}

\begin{figure}[!h]
    \centering
    \includegraphics[width=0.85\textwidth]{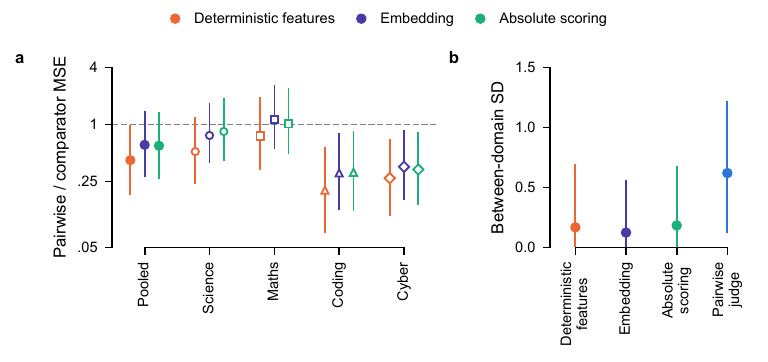}
    \caption{
        \textbf{Pairwise judges' prediction performance beats other protocols reliably in \code and \cyber}. \textbf{a,} Ratio of pairwise judging relative to each other protocol, pooled and by domain; values below one favour pairwise judging. \textbf{b,} Between-domain standard deviation $\sigma_p$ for each protocol on the log-MSE scale. Pairwise judging performance is also the most variable across domains. Error bars show 95\% credible intervals.
    }
    \label{fig:domain_glm_contrasts}
\end{figure}

In \code and \cyber, pairwise judging also outperforms every other protocol, with pairwise-to-comparator MSE ratios between 0.20 and 0.36 and all corresponding intervals below one (Figure~\ref{fig:domain_glm_contrasts}). Embeddings provide a smaller but more consistent reduction in error, with a pooled MSE ratio of 0.69 (0.52--0.95) and domain-specific ratios between 0.65 and 0.73. Absolute scoring improves slightly on the baseline in \science and \code (though less so than pairwise judging), while deterministic features show no clear improvement in any domain.

The advantage of pairwise judging being concentrated in \code and \cyber is consistent with this protocol working best in agentic tasks, but could also be related to subject matter. Additional data to directly compare agentic and non-agentic tasks within a domain is needed to disentangle these possibilities.

\subsection{Transfer of predictions across domains}
\label{app:domain-transfer}

We ask whether capability signals extracted from one domain's transcripts transfer to capability prediction in another domain. For each ordered pair of domains, we calibrate pairwise verdicts from the source domain against the reference models' measured capability in the target domain, while keeping the judge fixed to the best judge selected for the source domain.
We compare each cross-domain prediction with both the target domain's own transcripts and the baseline protocol.

\begin{figure}[!h]
    \centering
    \includegraphics[width=0.4\textwidth]{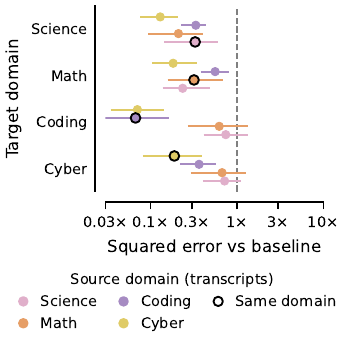}
    \caption{
        \textbf{Pairwise judging transfers asymmetrically across domains.}
        Squared prediction error relative to the saturated-item baseline when transcripts from one source domain are used to predict capability in another target domain. Outlined markers indicate within-domain predictions using transcripts from the target domain itself. Error bars show 95\% intervals; the dashed line marks equal error to the baseline.
    }
    \label{fig:domain-generalisation}
\end{figure}

Transcripts from the agentic domains transfer well to other domains (Figure~\ref{fig:domain-generalisation}). \code and \cyber transcripts predict each other's capability better than the baseline, with squared errors of 0.35 and 0.33 compared with baseline errors of 2.14 and 1.39 ($p\leq.001$). They also predict \maths and \science capability about as well as those domains' own transcripts, with cross-domain errors between 0.08 and 0.31 compared with within-domain errors of 0.42 and 0.21 ($p\geq.055$).

The reverse transfer is substantially weaker. \maths and \science transcripts predict \code capability with 18 and 11 times the within-domain error, respectively (2.52 and 1.59; $p\leq.015$), and predict \cyber capability with 4.4 and 3.6 times the within-domain error (1.16 and 0.94; $p\leq.025$). Neither transfer performs significantly better than the baseline ($p\geq.12$). Cross-domain transfer is therefore asymmetric: agentic transcripts retain capability information that transfers to non-agentic domains, while the reverse is not observed.

\subsection{Number of Evaluated Items}
\label{app:item_abl}

We next test how prediction performance changes with the number of saturated items available to each protocol. This measures whether the protocols benefit from additional transcript evidence or whether performance is primarily limited by systematic errors in the extracted signal.

For both absolute scoring and pairwise judging, additional items improve prediction in some settings but provide little benefit in others. In at least two cases (pairwise judging in \maths 2026-I and absolute scoring in \science 2025-I), performance worsens as more items are added and does not subsequently recover. This is consistent with systematic judging errors being reinforced as additional judgements accumulate, rather than averaged away.

\begin{figure}[!h]
    \centering
    \includegraphics[width=\linewidth]{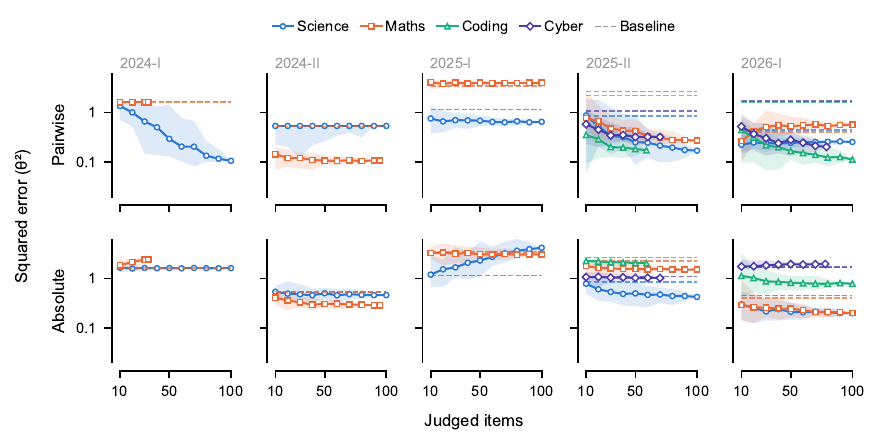}
    \caption{
    \textbf{Effect of the number of evaluated items.}
    Squared prediction error of pairwise judging and absolute scoring (the judge selected by reference leave-one-model-out error, as in the main analysis) as the number of judged saturated items varies across periods and domains: median over 30 random item subsets, band 2.5--97.5 percentile; \textit{dashed line} is baseline protocol error of each task.
    }
    \label{fig:item_abl}
\end{figure}

Overall, these results suggest that collecting more saturated transcripts does not necessarily resolve the main limitations of transcript-based capability estimation. For pairwise judging in particular, systematic judge bias can dominate sampling noise in some settings.

\subsection{Shifts in the Reference Population}
\label{app:reference_item_abl}
The prediction protocols are calibrated using a small set of reference models, so their performance may depend on which models are available for calibration. We test this sensitivity for pairwise judging by progressively restricting the reference population in two directions: retaining only the stronger reference models or retaining only the weaker reference models.

Removing the stronger references is substantially more damaging than removing the weaker ones. This suggests that high-capability references provide important anchors for locating target models on the capability scale, particularly when targets exceed much of the reference population.

\begin{figure}[!h]
    \centering
    \includegraphics[width=\linewidth]{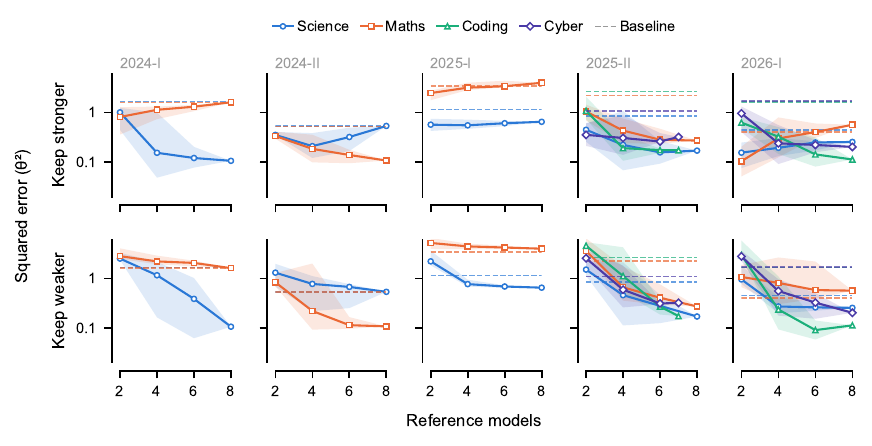}
    \caption{
    \textbf{Sensitivity to the reference population.}
    Squared prediction error of pairwise judging as the reference set is progressively restricted to stronger or weaker reference models: median over the third of reference subsets keeping the strongest (top) or weakest (bottom) models, band 2.5--97.5 percentile; \textit{dashed line}: the baseline protocol prediction for each task (all references).
    }
    \label{fig:reference_refcount_abl}
\end{figure}

Additionally, we tested whether excluding cases of self-judging (removing any cases where pairwise judges and absolute scoring judged their own performance) modified predictions.

\begin{figure}[h]
    \centering
    \includegraphics[width=.4\linewidth]{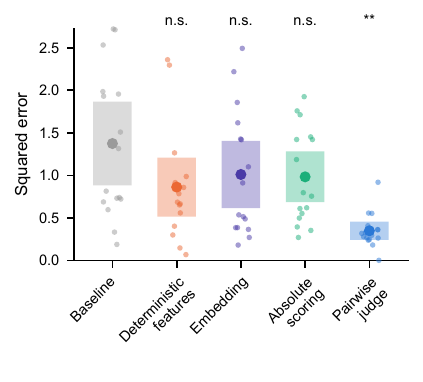}
    \caption{
    \textbf{Removal of self-judgements does not influence results.}
    Prediction performance of each protocol for 16 target models (squared error: lower is better), without self-judgements for absolute scoring and pairwise judging. This did not influence our results (cf.\ Figure~\ref{fig:signal-capability}). Each point is one target model; circle and box: mean and 95\% CI.
    }
    \label{fig:reference_abl}
\end{figure}

\subsection{Capability-Predicting Signal in Single Transcript Sections}
\label{app:thirds-kept-alone}

We complement the section-removal tests in Section~\ref{sec:results-source-position} with its converse: whether any third is \emph{sufficient}, by showing judges only the first, middle, or last third of every transcript.

\begin{figure}[!h]
    \centering
    \includegraphics[width=.85\textwidth]{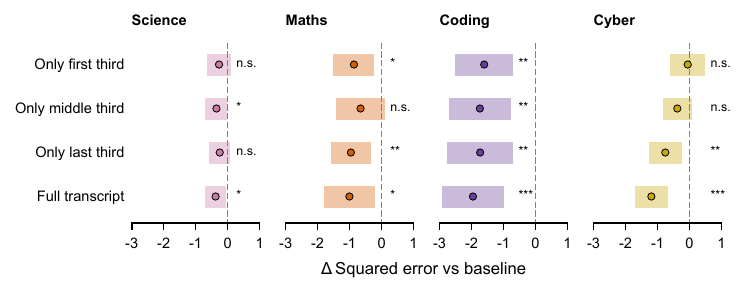}
    \caption{
    \textbf{Single transcript sections carry part of the capability information.}
        Change in prediction error of pairwise judging when only the first, middle, or last third of every transcript is shown, or the full transcript, relative to the baseline protocol (\textit{dashed line}). Points show mean changes across target models and boxes show 95\% paired $t$-intervals. *: $p<.05$; **: $p<.01$; ***: $p<.001$.
    }
    \label{fig:thirds-kept-alone}
\end{figure}

Overall, capability signals were distributed: shown any single third of transcripts, pairwise judging beat the baseline in most cases (Figure~\ref{fig:thirds-kept-alone}). In \code, any third was sufficient ($-1.61$ to $-1.74$, $p \leq .003$). In \maths, the first and last thirds were ($-0.86$, $p = .012$; $-0.96$, $p = .005$), and in \science only the middle third was ($-0.35$, $p = .048$).

In \cyber, however, capability-predicting information was concentrated towards the end of the transcript: only the final third outperformed the baseline ($-0.76$, $p = .007$); this is consistent with our results in Section~\ref{sec:results-source-position}, which show that removing the final third significantly degrades performance (Figure~\ref{fig:thirds}).

\subsection{Capability-Predicting Signal Across Transcript Sections For Other Protocols}
\label{app:thirds-other-protocols}

We repeat the transcript-position ablation from Section~\ref{sec:results-source-position} for the remaining transcript-based protocols. Each protocol is refitted after removing the first, middle, or final third of every transcript, to test whether the positional pattern observed for pairwise judging holds for alternative representations of transcript information.

\begin{figure}[H]
    \centering
    \includegraphics[width=.85\textwidth]{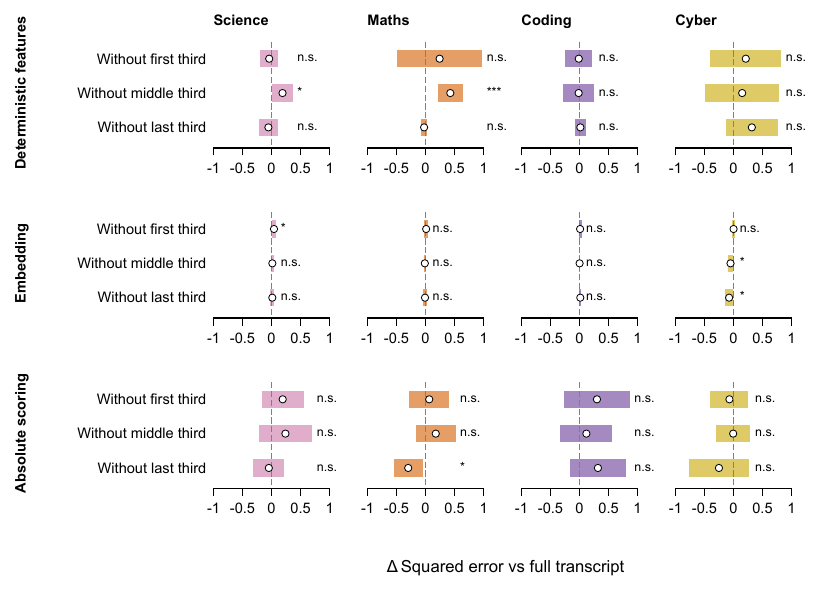}
    \vspace{-1em}
    \caption{
    \textbf{Capability signals across transcript positions for deterministic features, embeddings and absolute scoring.}
    Change in squared prediction error for deterministic features (\textit{top}), embeddings (\textit{middle}) and absolute scoring (\textit{bottom}) after removing the first, middle, or last third of every transcript, relative to the complete-transcript condition (\textit{dashed line}). Points show mean changes across target models and boxes show 95\% paired $t$-intervals. *: $p<.05$; ***: $p<.001$.
    }
    \vspace{-1em}
    \label{fig:thirds-other-protocols}
\end{figure}

As for pairwise judging, most removals left prediction error unchanged (Figure~\ref{fig:thirds-other-protocols}). For deterministic features, removing the middle third increased error slightly in \science ($+0.19$, $p = .037$) and \maths ($+0.43$, $p < .001$). Embedding error changed by at most $0.07$ in any domain: removing the first third increased it slightly in \science ($+0.04$, $p = .032$), and removing the middle ($-0.05$, $p = .012$) or last third ($-0.07$, $p = .044$) \textit{decreased} it in \cyber{} -- though note that the full-transcript error ($1.34$) is barely below that of the baseline protocol's prediction ($1.39$).
For absolute scoring, no removal significantly increased error; removing the last third \textit{decreased} it in \maths ($-0.30$, $p = .022$). No single third is consistently necessary for any of the three protocols across domains.

\subsection{Capability-Predicting Signal Across Transcript Channels For Other Protocols}
\label{app:channels-other-protocols}

\begin{figure}[H]
    \centering
    \vspace{-1em}
    \includegraphics[width=.85\textwidth]{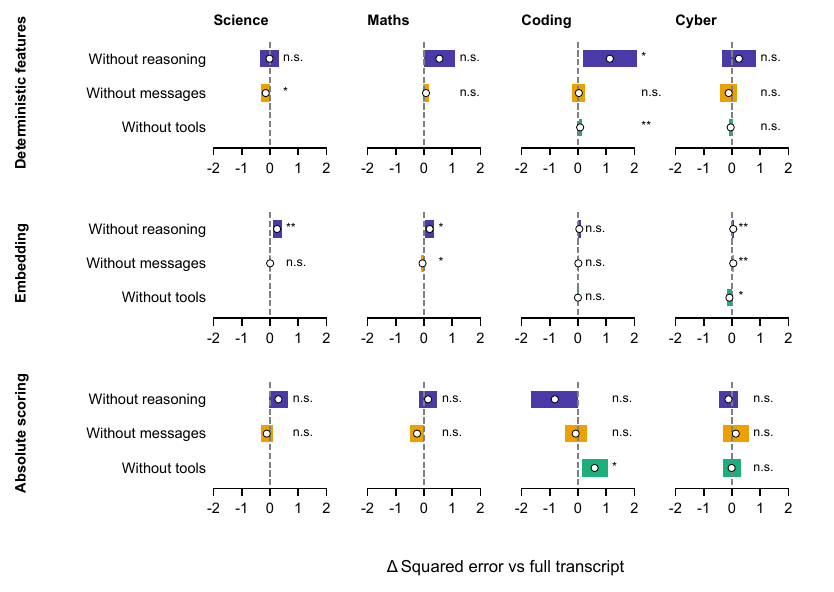}
    \vspace{-1em}
    \caption{
    \textbf{Capability signals across channels for deterministic features, embeddings and absolute scoring.}
    Change in squared prediction error for deterministic features (\textit{top}), embeddings (\textit{middle}) and absolute scoring (\textit{bottom}) after removing one transcript channel, relative to the complete-transcript condition (\textit{dashed line}). Deterministic features and embeddings are refitted using the corresponding channel-removed reference transcripts; for absolute scoring, the judge re-rates the channel-removed transcripts and is recalibrated on the same references. \maths and \science contain no tool calls. Points show mean changes across target models and boxes show 95\% paired $t$-intervals. *: $p<.05$; **: $p<.01$.
    }
    \label{fig:channels-other-protocols}
\end{figure}

We repeat the channel analysis from Section~\ref{sec:results-source-channels} for deterministic features, transcript embeddings, and absolute scoring to test whether the channels that matter for pairwise judging also carry signal for these alternative protocols. Each protocol is refitted after removing one channel from every transcript. Embeddings are recomputed from the channel-removed transcript, while deterministic features are recomputed from the remaining text and tool-use metadata.

We use the 16 target models from 2025-II and 2026-I in each domain. For each target, we compare squared prediction error under the channel-removed condition with error under the complete transcript using two-sided paired $t$-tests.

Unlike pairwise judging, no channel consistently carried signal for these protocols (Figure~\ref{fig:channels-other-protocols}). For deterministic features, removing reasoning increased error only in \code ($+1.13$, $p = .024$), and removing tools increased error slightly in \code ($+0.08$, $p = .006$). Removing reasoning increased embedding error in \science, \maths and \cyber ($+0.05$ to $+0.25$, $p \leq .012$), but all changes were small. For absolute scoring, only removing tools in \code significantly increased error ($+0.59$, $p = .015$).

\subsection{Results Are Robust to the Saturation Boundary}
\label{app:topk}

The main analysis defines an item as saturated when the three strongest reference models in a semester answer it correctly on their first attempt. Under this definition, weaker reference models and target models may still fail some saturated items. Since all protocols can only access correctly answered items, this can result in different models contributing slightly different subsets of transcripts. We therefore test whether the main results depend on this choice of saturation boundary.

We progressively tighten the definition. At level $k$, an item is considered saturated only if the $k$ strongest reference models answer it correctly, for $k=3,\ldots,8$. In the strictest ``all'' condition, every reference \textit{and} target model in the period must answer the item correctly. Tightening the boundary only removes items from the observed saturated set, so all analyses can be repeated using the existing transcripts, judge verdicts, ratings, features, and embeddings.

\begin{figure}[!h]
\centering
\includegraphics[width=0.62\linewidth]{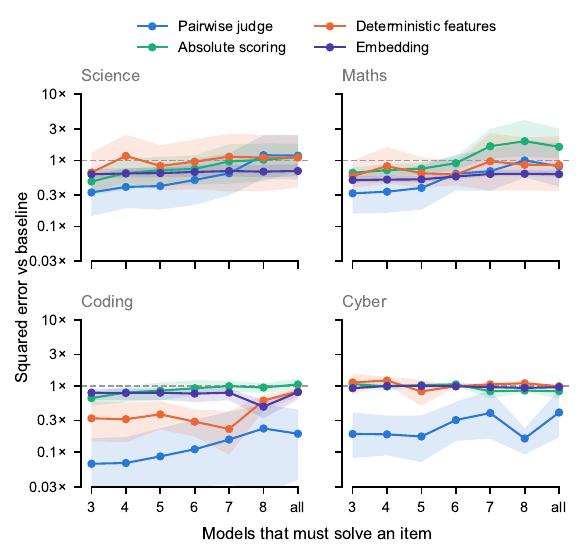}
\caption{
\textbf{Tightening the saturation boundary.}
Squared prediction error of each protocol relative to the baseline as the number of models required to solve an item increases. Levels 3--8 require the $k$ strongest reference models to solve the item; ``all'' requires every reference and target model in the period to solve it. $\theta^*$ is refitted at every level, and level 3 corresponds to the main-text setting. Bands show bootstrap 95\% intervals across the 16 target models from late 2025 and early 2026.
}
\label{fig:topk}
\end{figure}

At each level, we refit $\theta^*$ for the reference, target, and historical models using the corresponding unsaturated set while holding the global IRT item parameters fixed. The level-3 refit reproduces the frozen main-analysis capability estimates to within $0.001$. Because the unsaturated set changes with the boundary, the baseline prediction error also changes. We then refit and evaluate every protocol using the surviving observed items, selecting the judge independently at each level using leave-one-model-out error on the reference models. We report squared prediction error across the late-2025 and early-2026 target models relative to the corresponding baseline (Figure~\ref{fig:topk}; Table~\ref{tab:topk}).

\begin{table}[!h]
\centering
\small
\caption{
\textbf{Pairwise judging under stricter saturation boundaries.}
Squared prediction error relative to the corresponding baseline at each saturation level, with $\theta^*$ refitted separately at every level. Late 2025 and early 2026 are pooled. The reported $n$ values give the number of observed items summed across the two periods at level 3 and under the ``all'' condition.
}
\label{tab:topk}
\begin{tabular}{lrrrrrrr}
\toprule
& \multicolumn{7}{c}{Models that must solve an item} \\
\cmidrule(lr){2-8}
Target domain & 3 & 4 & 5 & 6 & 7 & 8 & all 16 \\
\midrule
Science ($n$: 200 $\to$ 140) & 0.33 & 0.40 & 0.41 & 0.51 & 0.64 & 1.20 & 1.18 \\
Maths ($n$: 200 $\to$ 102) & 0.32 & 0.34 & 0.38 & 0.63 & 0.69 & 1.00 & 0.81 \\
Coding ($n$: 159 $\to$ 52) & 0.07 & 0.07 & 0.09 & 0.11 & 0.16 & 0.23 & 0.19 \\
Cybersecurity ($n$: 149 $\to$ 94) & 0.19 & 0.19 & 0.17 & 0.31 & 0.39 & 0.16 & 0.40 \\
\midrule
Pooled & 0.19 & 0.20 & 0.22 & 0.32 & 0.39 & 0.65 & 0.49 \\
\bottomrule
\end{tabular}
\end{table}

Tightening the boundary progressively reduces the number of observed items, although the rate differs by domain. Across the two periods, \science retains 146 of 200 sampled items at $k=8$ and 140 under the ``all'' condition; \maths retains 109 and 102; \code retains 37 and 52; and \cyber retains 66 and 94. Note that for \code and \cyber, the 2025-II cohort contains only seven reference models, so the $k=8$ condition includes only 2026-I.

The refitted capability estimates change most for weaker target models. For example, the \science capability of gpt-oss-20b falls from 0.01 at level 3 to $-0.95$ at level 8, while the strongest target models move by less than 0.05. Baseline error therefore also decreases as the boundary tightens, from 0.65 to 0.51 in \science and from 1.32 to 0.97 in \maths.

Importantly, the ordering of protocols' performance is largely preserved. Pairwise judging has the lowest error at every boundary level in every domain except the strictest \science and \maths conditions, while absolute scoring, deterministic features, and embeddings remain closer to the baseline. However, the magnitude of the pairwise advantage decreases under stricter saturation criteria. Pooled across domains, pairwise error rises from $0.19\times$ baseline error at level 3 to $0.32\times$ at $k=6$, $0.65\times$ at $k=8$, and $0.49\times$ under the ``all'' condition.

The agentic domains are comparatively robust: pairwise judging changes from $0.07\times$ to $0.19\times$ baseline error in \code and from $0.19\times$ to $0.40\times$ in \cyber. In \maths, the ratio reaches $1.00\times$ at $k=8$ and $0.81\times$ under the ``all'' condition. \science is the main exception, increasing from $0.33\times$ at level 3 to $1.18\times$ under the strictest condition (95\% CI 0.51--2.41).

Two effects occur simultaneously as the boundary tightens. First, the observed saturated set loses items that weaker models fail, which can remove items on which differences between strong and weak correct solutions are easiest to detect. Second, those items move into the unsaturated set used to measure $\theta^*$, changing the capability target itself and generally compressing the target population towards the references. These effects are largest in \science, where tasks often produce short transcripts, and smallest in the agentic domains, where trajectories can differ substantially even when all models ultimately solve the same task.

\subsection{Predicting Across Longer Time Differences}
\label{app:static_pairwise}

In our main analyses, reference models and judges are always the models of the preceding period. This emulates a scenario where a newly released target model deviates only moderately from the references in its release date. Here, we ask how far these capability predictions can be stretched, by progressively increasing the distance between reference and target models. We sweep the reference period back from late 2025 to 2023, while keeping our target period fixed in early 2026.

\begin{figure}[!h]
    \centering
    \includegraphics[width=.8\linewidth]{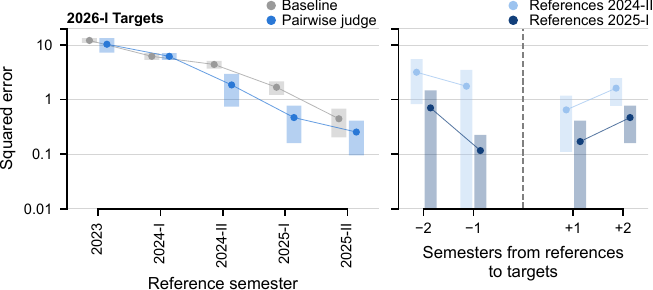}
    \caption{
    \textbf{Pairwise judging degrades with the distance between references and targets.}
    In \science, squared error per target; circle and box: mean and 95\% $t$-interval over the eight targets of a task.
    \textit{Left}: early 2026 targets predicted from each earlier reference period.
    \textit{Right}: references fixed at late 2024 (light) or early 2025 (dark), targets up to two periods older (negative) or newer (positive).
    Tasks one period ahead are those of the main text.
    }
    \label{fig:horizon}
\end{figure}

In \science, pairwise judging performance degrades as reference models get older. Prediction error rises from 0.25 with late 2025 references to 0.47 (early 2025) to 10.39 with 2023 reference models (Figure~\ref{fig:horizon}, left). It beats the baseline protocol with early 2025 ($p=.002$) and late 2024 references ($p<.001$), but not earlier references. Importantly, pairwise judging two periods back is worse than the baseline one period back.

This degradation in prediction could reflect the fact that future targets are far more capable than reference models, or simply a discrepancy between targets and references. To disentangle these possibilities, we fix the references at 2024-II or 2025-I and predict both older and newer targets, up to two periods away (Figure~\ref{fig:horizon}, right). We find that error grows with distance in both directions: targets two periods away were predicted worse than targets one period away (0.59 vs.\ 0.14, $p=.024$), while older and newer targets at the same distance did not differ ($p \geq .52$). Degradation is thus dependent on the distance between reference and target models, irrespective of direction.

\subsection{Cost of the Protocols}
\label{app:cost}
We price each protocol by the tokens its judge consumed, at prices listed on OpenRouter (September 2026).
For pairwise judging and absolute scoring, we trace costs as the number of judged items varies, similarly to Appendix~\ref{app:item_abl}.

The protocols differ in cost by orders of magnitude. Deterministic features are free and embeddings cost a few cents per semester, while absolute scoring costs \$2--7 and pairwise judging \$55--105, because every item is judged once per target--reference pair (quadratic number of comparisons).
Pairwise judging is the only protocol where additional spending generally buys accuracy. At the full item set, it reaches $0.07\times$ baseline error in \code, $0.19\times$ in \cyber, and $0.37\times$ in \science, and a tenth of the items (about \$6 per semester) already beats the baseline in the agentic domains.

\begin{figure}[!h]
    \centering
    \includegraphics[width=.6\linewidth]{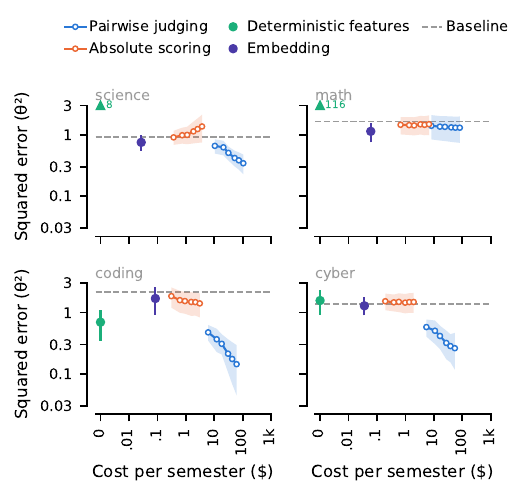}
    \caption{
    \textbf{Only pairwise judging turns cost into accuracy.}
    Prediction error against USD cost, pooled over every task of each domain and averaged over time periods. The judge is the reference with the lowest leave-one-out error. \textit{Dashed line}: baseline protocol. Pairwise judging and absolute scoring are shown as curves over 10--100\% of the judged items, with cost proportional to the items judged. Deterministic features and embeddings are single points.
    }
    \label{fig:cost}
\end{figure}

\subsection{Judge Ensemble}
\label{app:protocols:ensemble}
\begin{figure}[h]
    \centering
    \includegraphics[width=.6\linewidth]{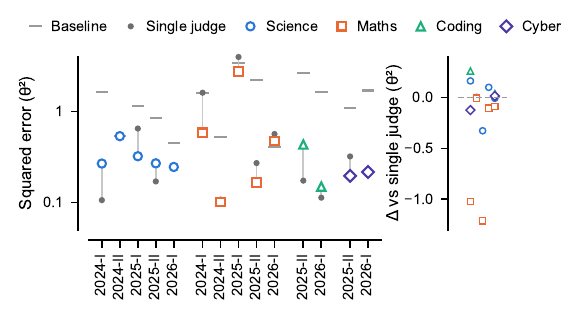}
    \caption{
    \textbf{Performance of the pairwise judge ensemble.}
    Squared prediction error of the ensemble of the three candidate judges (mean of their predictions; coloured markers) and of the judge selected by leave-one-model-out error (grey dots), on the items all three judged; grey tick, the baseline protocol. Right panel shows the paired difference (ensemble minus single judge; negative = ensemble is better). Ensemble error is generally similar to the single judge's, but substantially reduces the error of the selected judge in \maths 2024-I (1.61 to 0.58) and 2025-I (3.97 to 2.75).
    }
    \label{fig:judge_ensemble}
\end{figure}

We additionally evaluate a judge ensemble that averages the predictions of all three candidate judges for each prediction task (a single domain and semester). We compare this ensemble with the single judge selected using leave-one-model-out prediction error (Figure~\ref{fig:judge_ensemble}).

\clearpage

\end{document}